\documentclass{article}

\usepackage{microtype}
\usepackage{graphicx}
\usepackage[english]{babel}
\usepackage[utf8x]{inputenc}
\usepackage[T1]{fontenc}

\usepackage{booktabs} 
\usepackage{amsmath}
\usepackage{amsfonts}
\usepackage{caption}
\usepackage{subcaption}
\usepackage{todonotes}

\PassOptionsToPackage{numbers, compress}{natbib}

\usepackage{hyperref}

\usepackage{amsthm}

\newtheorem{proposition}{Proposition}[section]

\usepackage[final]{neurips_2022}
\usepackage{wrapfig}

\input m.macros
\input m.symbols
\usepackage{bbold} 

\usepackage{mbubble}
\mbubbleset{font=\tiny}

\newcommand\equaltoc{\mathrel{\overset{\makebox[0pt]{\mbox{\normalfont\tiny\sffamily $+ c$}}}{=}}}
\newcommand{\GP}{\mathcal{G}\mathcal{P}}

\def\qvi{q_{\text{VI}(\mathcal Q)}}
\def\Fvi{\mathcal{F}_{\text{VI}}}

\def\qsai{q_{\text{SRVAE}}}
\def\Fsai{\mathcal{F}_{\text{SRVAE}}}

\def\operator#1{\mathfrak{#1}}

\def\qsvae{q_{\text{SVAE}}}

\begin{document}

\title{Structured Recognition for\\ Generative Models with Explaining Away}

\author{%
Changmin Yu$^{1, 2}$ \quad Hugo Soulat$^{2}$ \quad Neil Burgess$^1$ \quad Maneesh Sahani$^2$ \\
$^1$Institute of Cognitive Neuroscience; \quad $^2$Gatsby Computational Neuroscience Unit; 
\\UCL, London, United Kingdom \\
\texttt{\{changmin.yu.19; hugo.soulat.19; n.burgess\}@ucl.ac.uk}; \\\texttt{maneesh@gatsby.ucl.ac.uk}
}

\maketitle

\begin{abstract}
  A key goal of unsupervised learning is to go beyond density estimation and sample generation to
  reveal the structure inherent within observed data.  Such structure can be expressed in the
  pattern of interactions between explanatory latent variables captured through a probabilistic
  graphical model.  Although the learning of structured graphical models has a long history, much
  recent work in unsupervised modelling has instead emphasised flexible deep-network-based
  generation, either transforming independent latent generators to model complex data or assuming
  that distinct observed variables are derived from different latent nodes.
  Here, we extend amortised variational inference to incorporate structured factors
  over multiple variables, able to capture the observation-induced posterior dependence between
  latents that results from ``explaining away'' and thus allow complex observations to depend on
  multiple nodes of a structured graph.
  We show that appropriately parametrised factors can be
  combined efficiently with variational message passing in rich graphical structures.
  We instantiate the framework in nonlinear Gaussian Process Factor Analysis, evaluating
  the structured recognition framework using synthetic data from known generative processes.
  We fit the
  GPFA model to high-dimensional neural spike data from the hippocampus of freely moving
  rodents, where the model successfully identifies latent signals that correlate with behavioural
  covariates.




\end{abstract}

\section{Introduction}
\label{sec: intro}

A central challenge of unsupervised learning is to identify and model patterns of statistical
dependence in high-dimensional data.
One approach to this challenge exploits latent-variable generative models, which capture the
statistical structure of data through the modelled influence of the latent variables on
observations, and through interactions between the latents themselves.
Recent developments in deep learning have enabled deep generative models (DGM), with remarkable
success in density estimation and high-fidelity image or text generation~\cite{kingma2013auto,
  rezende2014stochastic, goodfellow2014generative, gregor2015draw, oord2016conditional}.
However, much of the DGM development has emphasised the expressiveness and accuracy of the
latent-to-observation generative process, with the latents themselves often assumed independent
\textit{a priori}.

Variational learning requires inferring or approximating the posterior distribution of the latents
conditioned on observations~\cite{wainwright2008graphical}.
In a variational autoencoder~\cite[VAE;][]{kingma2013auto,rezende2014stochastic}, inference for a
DGM is \emph{amortised} by training a recognition network to return parameters of the
variational posterior.
The structured variational autoencoder~\cite[SVAE;][]{johnson2016composing} connects the DGM
framework with richer models that include structured prior dependence amongst the latent variables.
In an SVAE the structured latent prior is combined with a DGM link between each latent and a
corresponding observation.
Variational inference is amortised using recognition networks that return parameters of a factor
associated with each link, which are combined with the structured prior to obtain the full
posterior.
However, in many generative models of interest, observations depend on more than one latent
variable.
Such interactions induce joint potentials in the likelihood, coupling latents in the posterior even
if they are independent \textit{a priori}, a phenomenon sometimes called ``explaining
away''~\cite{pearl1988probabilistic}.
To capture such observation-induced dependence in the DGM setting we develop the \textit{structured
  recognition VAE} (SRVAE), where recognition potentials that incorporate joint factors induced by the
observations are learnt in a (structured) variational autoencoding framework.

We instantiate the SRVAE framework for a range of graphical structures, including non-linear latent Gaussian process (GP) models. In the latter case, we bring together
the GP-prior VAE~\cite{casale2018gaussian,ashman2020sparse}, and (sparse) Gaussian process factor
analysis (GPFA)~\cite{yu2008gaussian,duncker2018temporal} models, yielding a novel autoencoding
model (SR-nlGPFA).
Experiments show that SR-nlGPFA outperforms alternatives that lack structured recognition.  
We apply SR-nlGPFA to data collected from a population of neurons in the hippocampal complex, and
show that the unsupervised approach automatically captures dimensions underlying neural firing that
reflect relevant behavioural correlates.

\section{Background}

\label{sec: background}
\subsection{Variational Inference and Structured Variational Autoencoders}

Consider a general generative process with latent variables $z$ and observations $y$,
\begin{equation*}
    \begin{split}
        z\sim p(z|\theta), \quad y \sim p(y|z; \gamma) \,,
    \end{split}
\end{equation*}
where $\theta$ and $\gamma$ are the parameters of the prior and conditional likelihood
distributions, respectively (and we write $\Theta=\theta \cup \gamma$).
We assume deterministic $\theta$ for simplicity, but distributions over $\theta$ (with conjugate
hyperpriors) can be incorporated by variational Bayes (VB)~\cite{attias1999inferring}.
The posterior distribution, $p(z|y,\Theta)\propto p(z|\theta)p(y|z, \gamma)$, is often
analytically intractable.  In such cases, a common approach is to seek an approximation $q(z)$
constrained to a tractable class $\mathcal{Q}$ by variational inference (VI); that is, by minimising
the KL-divergence with respect to the true posterior distribution or, equivalently, maximising a
variational free energy~\cite{wainwright2008graphical} $\Fvi = \angles[\big]{\log p(y, z| \Theta)-\log q(z)}_{q(z)}$:
\begin{equation}
\begin{split}
  \qvi(z|y,\Theta) = \argmin_{q\in\mathcal{Q}}\KL{q(z)}{p(z|y, \Theta)} =
    \argmax_{q\in\mathcal{Q}}\angles[\big]{\log p(y, z| \Theta)-\log q(z)}_{q(z)}\,.
\end{split}
\label{eq: free_energy}
\end{equation}
In the variational autoencoder (VAE) architecture, the parametric optimisation of $q(z)$ implicit in
\eqref{eq: free_energy} is amortised by a recognition network that takes as input the observed
values $y$ and returns parameters of $q$.  The parameters of this recognition network are then
trained jointly with the generative model by stochastic optimisation of the free
energy~\cite{kingma2013auto, rezende2014stochastic}.

In both standard and amortised VI, the approximate distribution $q$ is often constrained to factor
over the latent variables, a so-called \emph{mean-field} constraint.  While tractable, such mean-field
approximations may be too restrictive to capture the complexity of the true
posterior~\cite{saul1995exploiting, hoffman2015structured}.
Many approaches have been proposed for improving the expressiveness of variational
approximation~\cite{blei2006variational, tran2015variational, li2017approximate, lee2020meta}. Here
we consider the structured VAE~\cite[SVAE;][]{johnson2016composing}.
Unlike standard VAEs, the SVAE assumes the generative prior distribution to be specified by a
structured probabilistic graphical model (PGM), $p(z|\theta) \propto \prod_{c\in C}\psi_{c}(z_{c})$,
where $\{\psi_c\}_{c\in C}$ correspond to $C$ clique potentials.
In addition, the amortised inference network outputs recognition factors, $r(z|y, \phi)$, that
approximate the generative likelihood function rather than the full variational posterior.  These
recognition factors are combined with the structured prior distribution to obtain the amortised
variational posterior,
\begin{equation}
  \qsvae(z|y, \Theta, \phi) =
    \argmax_{q \propto r(z|y, \phi)p(z|\theta)}\sum_{y\in\mathcal Y}\angles{\log\lr(){\frac{p(z|\theta)p(y|z, \gamma)}{q(z)}}}_{q(z)}
    \propto r(z|y, \phi^*)p(z|\theta)\,,
    \label{eq: svae_inference}
\end{equation}
where $\phi^*$ minimises the averaged KL over the data observations $\mathcal{Y}$.
\citet{johnson2016composing} considered recognition factors that are local (singleton) evidence potentials, 
chosen to be conjugate to $p(z|\theta)$.
Even in this case, the form of \eqref{eq: svae_inference} allows the dependency structure established
by the prior to be carried over to the variational posterior distribution. Further
details of SVAE appear in Appendix~\ref{sec: app_svae_further}.

\subsection{Gaussian Process Factor Analysis}
Gaussian Process Factor Analysis (GPFA) is a model used in neural data analysis to infer the dynamic
latent structure underlying high-dimensional population spike trains~\cite{yu2008gaussian,duncker2018temporal}.
Standard GPFA assumes the following generative model:
\begin{equation}
\begin{split}
    \text{latent functions: } f^{k}(\cdot) &\sim\GP\lr[\big](){m_{\theta}^{k}(\cdot), \kappa_{\theta}^{k}(\cdot, \cdot)}, \text{ for } k = 1, \dots, K\,, \\
    \text{affine embeddings: } h_{n}(\cdot) &= \sum_{k=1}^{K}c_{nk}f^{k}(\cdot) + d_{n}, \text{ for } n = 1, \dots, N\,, \\
    \text{observations: } y_{n}(t) &\sim p\lr[\big](){y_{n}(t)|g(h_{n}(t))}, \text{ for } t = 1, \dots, T\,,
\end{split}
\label{eq: gpfa_gen}
\end{equation}
where $m_{\theta}^{k}(\cdot)$ and $\kappa_{\theta}^{k}(\cdot, \cdot)$ parametrise the GP prior,
$c_{jk}$ and $d_{n}$ define an affine mapping from the latent space to the observation space (or a
transform of the observations for general likelihoods), and $g(\cdot)$ is a smooth scalar link
function appropriate for the observation distribution.
We take $m^{k}_{\theta}(\cdot) = 0$ unless stated otherwise.  Correlations between observed neurons
are captured by dependence on the common set of latents, while temporal correlations in the
high-dimensional observations are modelled by the latent temporal correlations of the GPs.  The PGM
describing the GPFA generative model (with sparse variational approximation, see below) is shown in
Fig~\ref{fig: svgpfa_generative}.

Although maximum-likelihood parameters for GPFA can be found using (variational)
expectation-maximisation~\cite{dempster1977maximum, yu2008gaussian} exact GP inference scales
cubically in the number of observation times~\citep{rasmussen2005gaussian}.  Sparse variational GP
(svGP) inference based on auxilliary inducing points~\cite{titsias2009variational} reduces the time
complexity of learning in GPFA~\cite{adam2016scalable}, and also facilitates efficient extensions to
(pointwise) non-conjugate likelihoods~\cite{duncker2018temporal}.

For $k=1, \dots, K$, we introduce inducing values $\vec{u}_{k}$ representing the evaluations of the
latent process $f^{k}$ at $M_{k}$ inducing locations, $\vec{z}_{k}$. For simplicity, we assume
$M_{1} = \cdots = M_{K} = M$ unless stated otherwise. The GPFA generative model \eqref{eq: gpfa_gen}
can be augmented to include these auxiliary variables:
\begin{equation}
\begin{split}
    p(\vec{u}_{k}|\vec{z}_{k}) = \normal{\vec{0}}{\vec{K}_{\vec{z}_{k}\vec{z}_{k}}^{k}}, \quad
    p(f^{k}(\cdot)|\vec{u}_{k}) = \GP( \operator{F}_{k}(\cdot)\vec{u}_{k}, \kappa_{\theta}^{k}(\cdot,
    \cdot) - \operator{F}_{k}(\cdot)\vec{K}_{\vec{z}_{k}\vec{z}_{k}}^{k} \operator{F}_{k}^T(\cdot)),
\end{split}
\end{equation}
where $\operator{F}_{k}(\cdot) = \kappa^{k}_{\theta}(\cdot,
\vec{z}_{k})(\vec{K}^{k}_{\vec{z}_{k}\vec{z}_{k}})^{-1}$ is the linear operator that maps
$\vec{u}_{k}$ to $f^{k}(\cdot)$; $\kappa^{k}_{\theta}(\cdot, \vec{z}_{k})$ is a vector-valued
function with $\kappa^{k}_{\theta}(x; \vec{z}_{k}) = [\kappa^{k}_{\theta}(x, z_{k1}),
  \kappa^{k}_{\theta}(x, z_{k2}), \dots, \kappa^{k}_{\theta}(x, z_{kM_{k}})]$; and
$\vec{K}_{\vec{z}_{k}\vec{z}_{k}}^{k}$ is the covariance matrix obtained by evaluation of the kernel
function $\kappa^{k}_{\theta}(\cdot)$ at the inducing locations $\vec{z}_{k}$.

Introducing a variational distribution over the inducing points and the latent functions, and
utilising the generative model (Fig~\ref{fig: svgpfa_generative}), the free energy of the sparse
variational GPFA (svGPFA) can be expressed as following: 
\begin{equation}
  \mathcal{F}(\Theta, \vec{C}, \vec{d}, \vec{Z}, \phi) 
  = \angles{\log
    \frac{\lr(){\prod_{t=1}^{T}
        p(\vec{y}_{t}|\vec{f}_{t}; \gamma, \vec{C}, \vec{d})
        }
        p(\vec{F}|\vec{U}; \theta)
        p(\vec{U}|\vec{Z}; \theta)
    }{
        q(\vec{F}, \vec{U}; \phi)
    }
}_{q(\vec{F}, \vec{U}; \phi)} \,.
  \label{eq: svgpfa_free_energy}
\end{equation}
where $\vec y_t = [y_n(t)] \in\reals^N$, 
$\vec f_t = [f^k(t)] \in\reals^K$, $\vec F = [\vec f_{t}]^T \in \reals^{T\times K}$, $\vec C = [c_{nk}] \in\mathbb{R}^{N\times K}$, $\vec d = [d_{n}] \in\mathbb{R}^{n}$, $\vec U = [\vec u_{1}, \dots, \vec u_{K}]^T\in \mathbb{R}^{K\times M}$, and $\vec Z = [\vec z_{1}, \dots, \vec z_{K}]^T\in\mathbb{R}^{K\times M}$.
The svGP approach constrains $q(\vec{F}, \vec{U}; \phi)$ to the form $q(\vec{U})
p(\vec{F} | \vec{U}, \theta)$ with Gaussian $q(\vec{U})$~\cite{titsias2009variational}.  In
the multi-GP case, the distribution is typically also taken to factorise over
processes~\cite{adam2016scalable}, $q(\vec{F}, \vec{U}) =
\prod_{k=1}^{K}p(\vec{f}^{k}|\vec{u}_{k})q(\vec{u}_{k})$, with $\vec{f}^{k} = [f^{k}(1), \dots, f^{k}(T)] \in \reals^{t}$, $q(\vec{u}_{k}) =
\mathcal{N}(\vec{m}_{k}, \vec{S}_{k})$.  Under these constraints on $q$, the svGPFA free energy
simplifies to
\begin{equation*}
    \mathcal{F}_{\text{svGPFA}} = \sum_{t} \angles{\log p(\vec{y}_{t}|\vec{h}_{t})}_{q(\vec{h}_{t})} -
    \sum_{k=1}^{K}\KL{q(\vec{u}_{k})}{p(\vec{u}_{k}|\vec{z}_{k})} \,,
\end{equation*}
where $q(\vec{h}_{t}) = \intdx[d\vec{U}]{p(\vec{h}_{t}|\vec{U}) q(\vec{U})}$  is
itself a GP with mean and kernel functions given by
\begin{equation*}
    \begin{split}
        m^{h}_{n}(t) = \sum_{k=1}^{K}c_{nk}\operator{F}_{k}(t)\vec{m}_{k} + d_{n} \text{ and }
        \nu^{h}_{n}(t, t') = \sum_{k=1}^{K}c_{nk}^2(\kappa_{k}(t, t') +
        \operator{F}_{k}(t)(\vec{S}_{k}-\vec{K}_{\vec{z}_{k}\vec{z}_{k}})\operator{F}_{k}(t')^T) \,.
    \end{split}
\end{equation*}
Hence the computational cost for sparse variational GP inference reduces from $\mathcal{O}(T^3)$ down to $\mathcal{O}(M^3 + TM^2)$. 

\section{Structured Recognition and Explaining Away}
\label{sec: method}

\subsection{Structured Recognition in VAEs}

A general non-linear likelihood may induce additional latent dependency structure in the posterior
beyond that of the prior (a simple illustrative example appears in Appendix~\ref{sec:
  app_tree_joint_factor}).
Although the SVAE approach developed by \citet{johnson2016composing} composed amortised inference
with a structured prior distribution, the recognition models these authors discussed all contributed
amortised potentials that factored over latent variables.
Such an approach cannot accurately model posteriors with dependence structure that differs from that 
of the prior PGM.

Here we adopt a recognition network that outputs structured factor potentials over the latents,
providing amortised estimates of the additional dependencies induced by ``explaining away''.  We
denote the resulting model the \textit{Structured Recognition VAE} (SRVAE).  The SRVAE variational
approximation takes the form
\begin{equation}
    \qsai(z|y, \theta, \phi) \propto \prod_{c\in C}\psi_{c}(z_{c};\theta)\prod_{c\in C_{r}}\xi_{c}(z_{c}|y; \phi)\,,
    \label{eq: aea_e_step}
\end{equation}
where $C_r$ is the set of recognition factors and the $\xi_{c}$s the associated factor potentials.
We assume that the $\xi_c$s are chosen to be conjugate to the prior factors unless otherwise stated. 
Hence the analytical (approximate, if necessary) form of $\qsai$ can be computed with (variational)
message passing.  In the most general form, we could let $\xi(z|y; \phi)$ be a single joint factor
potential over the complete set of latent variables.

The free energy objective takes the same form as the standard autoencoding free energy
objective~\cite{kingma2013auto, rezende2014stochastic}, but with the structured amortised
variational approximation, $\qsai$, as the variational distribution.
\begin{equation}
    \Fsai(\Theta, \phi) = \langle\log p(z|\theta) + \log p(y|z, \gamma) - \log \qsai(z|y, \phi, \theta)\rangle_{\qsai(z|y, \phi, \theta)}
    \label{eq: aea_free_energy}
\end{equation}
The training of the model follows standard VAE-style stochastic optimisation to update the
parameters of the recognition and generative networks, as well as (optionally) the parameters of the
prior PGM.

The following proposition supports the use of SRVAE framework (the proof
appears in Appendix~\ref{sec: app_prop}).
\begin{proposition}
\label{sec: aea_lower_bound}
The SRVAE objective function provides a tighter lower bound to the free energy than the SVAE objective function. 
\begin{equation}
    \max_{q}\mathcal{F_{\text{VI}(\mathcal Q)}}(\Theta,q) \geq \max_{\phi} \Fsai(\Theta, \phi) \geq \max_{\phi} \mathcal{F}_{\text{SVAE}}(\Theta, \phi)
    \label{eq: aea_lower_bound}
\end{equation}
\end{proposition}

Below we describe two instantiations of the SRVAE framework, the first based on a latent Gaussian
mixture model, and the second on a latent GPFA model, both with DGM outputs. Structured amortised
inference facilitates scalable inference of the posterior latent distribution with full covariance
structure, allowing more accurate learning than with factored recognition approaches.
However, the SRVAE framework is more general-purposed, and can be combined with many different
latent variable models.  Further examples appear in Appendix~\ref{sec: app_prop}.

\begin{figure}[t!]
     \centering
     \begin{subfigure}[b]{0.25\textwidth}
         \centering
         \includegraphics[width=\textwidth]{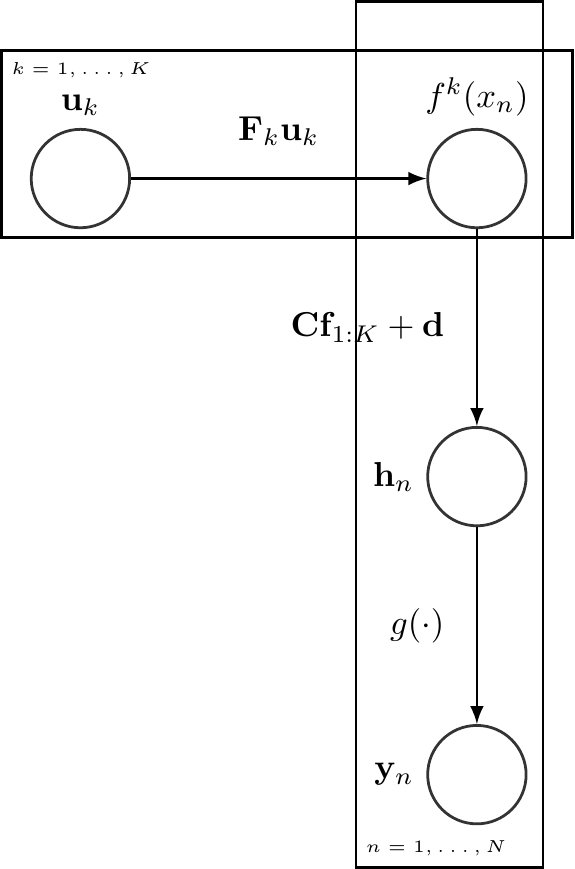}
         \caption{}
         \label{fig: svgpfa_generative}
     \end{subfigure}
     \hfill
     \begin{subfigure}[b]{0.3\textwidth}
         \centering
         \includegraphics[width=\textwidth]{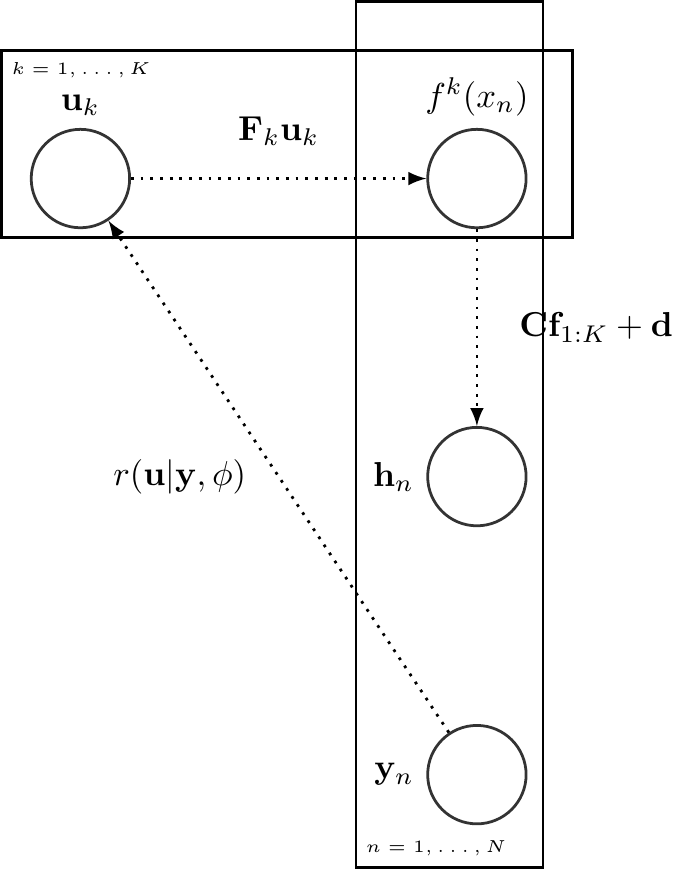}
         \caption{}
         \label{fig: svgpfa_inference}
     \end{subfigure}
     \hfill
     \begin{subfigure}[b]{0.3\textwidth}
         \centering
         \includegraphics[width=\textwidth]{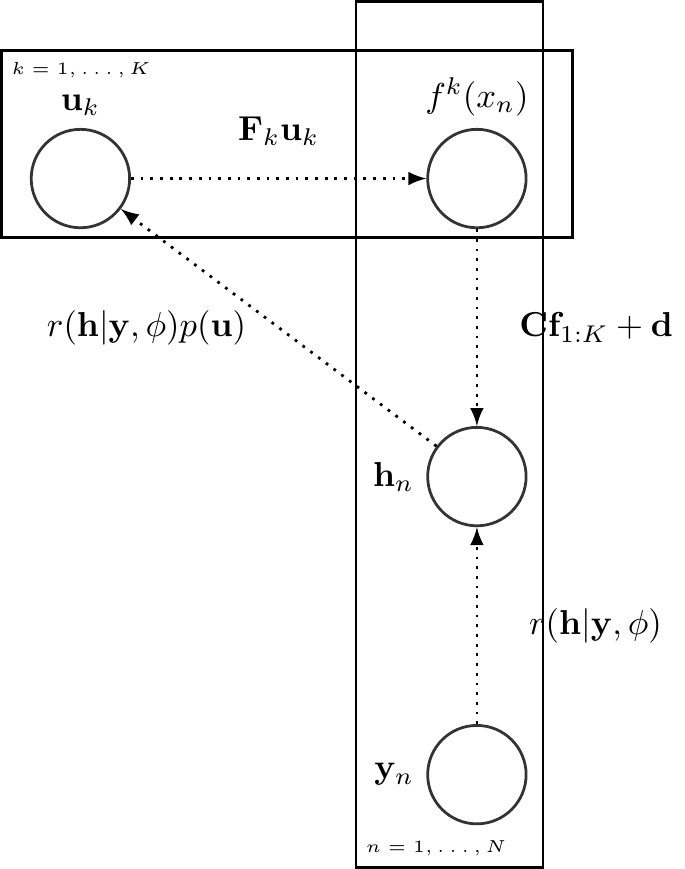}
         \caption{}
         \label{fig: aea_svgpfa_inference}
     \end{subfigure}
        \caption{\textbf{Graphical models for sparse variational GPFA models}. \textbf{(a)} Generative model of GPFA model with sparse inducing point approximation; \textbf{(b)} Standard GPFA inference with sparse amortised variational approximation (note that $\vec{C} = \vec{I}$ and $\vec{d} = \vec{0}$ for SGP-VAE~\cite{ashman2020sparse}); \textbf{(c)} Structured recognition potential enables full-covariance variational inference in SR-nlGPFA model.}
        \label{fig: inference_pgms}
\end{figure}

\subsection{Structured Recognition with Latent Gaussian Mixture Model}

We instantiate SRVAE with a classical latent variable model, the latent Gaussian mixture model (GMM)~\cite{bishop2006pattern}. We assume the following generative process (Figure~\ref{fig: gmm_generative}):
\begin{equation}
    \begin{split}
        z_{t}|\pi&\sim \text{Categorical}(z|\pi)\\
        \vec{h}_{t} &\sim \mathcal{N}(\vec{h}|\vec{\mu}^{(z_{t})}, \vec{\Sigma}^{(z_{t})}) \\
        \vec{y}_{t}|\vec{h}_{t} &\sim \mathcal{N}(\vec{y}|\vec{\mu}_{\text{NN}}(\vec{h}_{t}), \vec{\Sigma}_{\text{NN}}(\vec{h}_{t}))
    \end{split}
\end{equation}
This model was studied by \citet{johnson2016composing}, using fully factorised recognition
potentials, followed by variational message passing\cite[VMP;][]{winn2005variational} to obtain the
GMM variational posterior (details can be found in Appendix~\ref{sec: app_svae_further}).
However, the coupled conditional model, parametrised by a neural network, induces posterior
correlations between the latent variables that do not exist \textit{a priori}.
Hence in order to implement the structured amortised inference, we assume a full covariance Gaussian
recognition potential on $\vec h$,
and the SRVAE-GMM variational approximation is
computed with VMP given the recognition and prior factors.

\subsection{Structured Recognition Variational Autoencoding Nonlinear GPFA}
\label{sec: aea_svgpfa}

The GPFA model \eqref{eq: gpfa_gen} incorporates a generalised-linear likelihood, with the link
function $g(\cdot)$ acting separately on each affine embedding value $h_n$.  We are now in a position to
use the SRVAE framework to extend GPFA to include a DGM likelihood, greatly increasing the
expressiveness of the generative model.  Consider observations $\{(x_{t}, \vec{y}_{t})\}_{t=1}^{T}$,
defining $\vec{f}_t \in \reals^K$ and $\vec{h}_t\in\reals^N$ to be the corresponding vectors of
latent process values and embeddings at $x_t$. [In neural applications, inputs $x_t$ are usually taken to be the timestamp $t$; see \eqref{eq: gpfa_gen}.  Here we consider the more general case of arbitrary inputs.]
In the generative model, we retain the affine mapping from $\vec f_t$ to $\vec h_t$,
replacing the link function by a non-linear multivariate DGM: $\vec y_t \sim p(\vec y
| g(\vec h_t, \gamma))$.  Although the generative affine mapping could, in principle, be subsumed
within the general deep network, the embeddings $\vec{h}_t$ will play a valuable role in
parametrising structured recognition.

For amortised svGP inference, our goal is to define a variational distribution $q(\vec U
| \vec Y, \phi)$.  \citet{ashman2020sparse} have considered a similar latent GP model called the
SGP-VAE.  There, the variational distribution over the latents was found by combining the prior GP
distribution with an amortised approximation of the likelihood function factored over latent
processes~\cite{titsias2009variational}:
\begin{equation}
    q(\vec{F}, \vec{U}) = \prod_{k}p(\vec f^{k}|\vec{u}_{k})q(\vec{u}_{k}), \text{ where } q(\vec{u}_{k})\propto p(\vec{u}_{k})\prod_{t}r(\vec{u}_{k}|\vec{y}_{t}, x_{t}, \vec{z}_{k}),
\end{equation}
The graphical model for SGP-VAE inference corresponds to that of Figure~\ref{fig: svgpfa_inference}, with
$\vec{C}=\vec{I}$ and $\vec{d} = \vec{0}$. While this approach captures induced correlation amongst the
inducing points, it fails to capture correlations \emph{between} latent processes that arise in
the posterior through ``explaining away", potentially leading to sub-optimal inference and learning.

Our solution is to recast the output of the amortised inference to the likelihood of $\vec{h}$
rather than $\vec{U}$, and propose the following structured variational distribution. 
\begin{equation}
    q(\vec{F}, \vec{U}) = \left[\prod_{k}p(\vec f^{k}|\vec{u}_{k})\right]q(\vec{U}), \text{ where
    } q(\vec{U})\propto \int d\vec{H} \, p(\vec{U})p(\vec{H}|\vec{U})\prod_{t}r(\vec{h}_{t}|\vec{y}_{t}) 
    \label{eq: q_joint_aea}
\end{equation}
where $r(\vec{h}_{t}|\vec{y}_{t}) \propto \normal{\vec{h}_t|\vec{m}(\vec{y}_{t}; \phi)}{\vec{\Psi}(\vec{y}_{t}; \phi)} = \normal{\vec{\mu}_{t}^h}{\vec{\Psi}_{t}^h}$, with $\vec{\Psi}_{t}^{h}$ diagonal, 
and $\vec H = [\vec h(x_{1}), \dots, \vec h(x_{T})]$. 
Given the linear-Gaussian relationship between $\vec{U}$ and $\vec{h}$ (Fig~\ref{fig:
  svgpfa_generative}), the integral in \eqref{eq: q_joint_aea} can be computed in closed
form. Furthermore, even though the recognition potential on $\vec{h}$, $r(\vec{h}|\vec{y}_{t})$, is
assumed to be fully factorised over the dimensions of each $\vec{h}_t$, the variational distribution
on $\vec{U}$ includes coupling between latent processes induced by the affine mixing coefficients.
Hence $q(\vec{U})$ captures the correlations both between the latent processes and through time
(through combination with the GP prior under the structured autoencoding formulation).

Specifically, given the general expression of the GPFA generative model and the sparse approximation
with inducing points, the linear Gaussian relationship between $\vec{h}$ and $\vec{U}$ at any $x$ is
given by
\begin{equation}
    p(\vec{h}|\vec{U}, x) = \normal{\vec{C}\operator{F}(x)\vec{U} + \vec{d}}{\vec{C}(\vec{K}_{x} - \operator{F}(x)\vec{K}_{\vec{U}}\operator{F}(x)^T)\vec{C}^T}\,, 
    \label{eq: aea_svgpfa_ph}
\end{equation}
where
\begin{equation*}
    \operator{F}(x) = \begin{bmatrix} 
                       \operator{F}_{1}(x) & &  \\
                        & \ddots & \\
                        & & \operator{F}_{K}(x)
                      \end{bmatrix}, \quad 
    \vec{K}_{\vec{U}} = \begin{bmatrix} 
                       \vec{K}_{\vec{z}_{1}\vec{z}_{1}} & & \\
                        & \ddots & \\
                        & & \vec{K}_{\vec{z}_{K}\vec{z}_{K}}
                      \end{bmatrix}
\end{equation*}
with
$\vec{K}_{x} = \text{diag}[\kappa^{k}_{\theta}(x, x)]$. 
Combining this result with the fully factorised recognition potential on $\vec{h}$ at $x_{t}$ \eqref{eq:
  q_joint_aea}, we obtain the structured variational distribution on $\vec{U}$:
\begin{align}
\label{eq: svgpfa_qu}
    &q(\vec{U}) = \mathcal{N}(\vec{m}_{\vec{U}}, \vec{S}_{\vec{U}}) \propto p(\vec{U})\prod_{t}\mathcal{N}(\vec{C}\operator{F}(x_{t})\vec{U}|\vec{\mu}_{t}^h, \vec{\Psi}_{t}^h), \text{ with } \tag{13}\\
    & \vec{S}_{\vec{U}}^{-1} = \vec{K}_{\vec{U}}^{-1} + \sum_{t}\operator{F}(x_{t})^T\vec{C}^T(\vec{\Psi}^{h}_{t})^{-1}\vec{C}\vec{F}(x_{t}), \quad \vec{m}_{\vec{U}} = \vec{S}_{\vec{U}}\lr(){\sum_{t}\operator{F}(x_{t})^T\vec{C}^T(\vec{\Psi}_{t}^{h})^{-1}(\vec{\mu}^{h}_{t} - \vec{d})}\,. \notag
\end{align}
This variational posterior on $\vec{U}$ leads to a corresponding posterior on $\vec h$ at each $x$: 
\begin{equation*}
    q(\vec{h}(x)) = \intdx[d\vec U] p(\vec{h}|\vec{U})q(\vec{U}) = \normal{\vec{C}\operator{F}(x)\vec{m}_{\vec{U}} + \vec{d}}
        {\vec{C}(\vec{K}_{n} + \operator{F}(x)(\vec{S}_{\vec{U}}-\vec{K}_{\vec{U}})\operator{F}(x)^T)\vec{C}^T)}.
\end{equation*}
and we write $q(\vec h(x_t)) = \normal{\vec{m}^{h}_{t}}{\vec{S}^{h}_{t}}$.
Then the complete (reparametrised) Monte Carlo estimate of the free energy objective given a mini-batch of data, $\{(x_{b}, \vec{y}_{b})\}_{b=1}^{B}$, takes the following expression.
\begin{equation}
    \mathcal{F}(\theta, \gamma, \phi, \vec{C}, \vec{d}, \vec{Z}) = \sum_{b=1}^{B}\frac{1}{S}\sum_{s=1}^{S}\log p(\vec{y}_{b}|\vec{m}^{h}_{b} + \vec{L}^{h}_{b}\epsilon_{s}) - \KL{q(\vec{U})}{p(\vec{U})}
\end{equation}
where $\epsilon_{s}\sim\normal{\vec{0}}{\vec{I}}$, $\theta$ is the set of kernel parameters, $\vec{L}^{h}_{b}$ is the lower-triangular Cholesky component of $\vec{S}^{h}_{b}$ such that $\vec{S}^{h}_{b} = \vec{L}^{h}_{b}(\vec{L}^{h}_{b})^T$.

We note that we have retained the scalability of amortised inference by choosing an amortised
diagonal-Gaussian potential on $\vec{h}$.  However, the linear-Gaussian relationship between the
inducing points $\vec{U}$ and $\vec{h}(x)$ \eqref{eq: aea_svgpfa_ph} leads to a full-covariance
variational Gaussian approximation for $\vec{U}$, allowing amortised inference to capture the
observation-induced posterior correlations between latent processes (commonly referred to as the
``explaining away'' effect).
Thus, the retention of the affine GPFA mapping introduces a key extenstion to SGP-VAE.  We will
refer to the new model as \textit{structured recognition non-linear GPFA} (SR-nlGPFA).  The generative model corresponds to that of svGPFA
(Figure~\ref{fig: svgpfa_generative}) but with the nonlinearity $g(\cdot)$ generalised to a flexible
form modelled by a neural network. The complete inference procedure for SR-nlGPFA
is graphically illustrated in Figure~\ref{fig: aea_svgpfa_inference}.

The implicit computation of the variational distribution over $\vec{U}$ given the recognition
potentials on $\vec{h}$ also makes it possible to carry out svGP inference with changed inducing
locations
~\eqref{eq: svgpfa_qu}.
This is particularly useful in situations that require inference over test datasets of different
durations to those seen in training.
In particular, it may be possible to learn an amortised inference network using short sub-sequences
drawn from a longer dataset, and then infer a posterior over latent GPs for the complete data
sequence efficiently by optimising the placement of inducing points along its full length.
See further details on the free-form svGP inference step in Appendix~\ref{sec: app_aea_svgpfa}.

\section{Related work}
\label{sec: related_works}

\textbf{Structured Deep Generative Models.} A number of studies have considered latent graphical
structure within  the DGM framework~\cite{krishnan2015deep, dilokthanakul2016deep, johnson2016composing,
  lin2018variational, li2018graphical, paulus2020gradient}.
Our work builds on and generalises the prominent SVAE proposal of \citet{johnson2016composing}.
Both SVAE and SRVAE combine the structure of a prior PGM with the flexibility of neural
network-based recognition. The difference lies in the form of the recognition potentials.  These
were taken to extend over single latent variables in the earlier study.  Here, we consider joint
potentials that capture dependence induced by the generative likelihood.
In a closely related study, \citet{lin2018variational} proposed a structured inference network,
which approximates the variational distribution as the combination of the recognition potential and
a separate structured latent distribution independent of the prior, the independent structured
latent distribution can have non-conjugate factors to improve the expressiveness of the model.

\textbf{svGPFA and Extensions.}  Since its introduction as a model to identify linear
low-dimensional struction in neural population data~\cite{yu2008gaussian}, GPFA has been extended to
incorporate non-linear link functions and non-conjugate (notably Poisson count or point-process)
noise models, and combined with sparse variational inference for efficiency~\cite{adam2016scalable,
  zhao+al:2017:neuralcomp, duncker2018temporal, keeley2020efficient}.  
Related work has considered GP latents in the context of DGMs, particularly using variational
autoencoding~\cite{casale2018gaussian, campbell2020tvgp, ashman2020sparse}.
In a sense, SR-nlGPFA combines both approaches.  It incorporates multiple latent GPs (as in GPFA and
some DGM GP models) with an affine mapping that feeds into a nonlinear DGM.  However, this structure
means that the affine map contributes little to the generative process.  Instead it provides a
target for amortised inference, which combines with sparse variational GP inference to yield a full
structured posterior on inducing point values.
As such, the affine map may is better seen as an element of the structured recognition model.

\section{Results}
\label{sec: results}
We evaluated SRVAE methods for both latent mixtures, and nlGPFA, comparing to relevant baselines on
synthetic and real datasets.  Empirical results on the instantiations of the SRVAE framework with
other latent variable models can be found in Appendix~\ref{sec: app_further_results}\footnote{Python
  implementation can be found at
  \url{https://github.com/changmin-yu/structured-recognition-neurips-2022}}.

\subsection{Experiments with SRVAE-GMM}
\begin{figure}[t]
     \centering
     \begin{subfigure}[b]{0.32\textwidth}
         \centering
         \includegraphics[width=\textwidth]{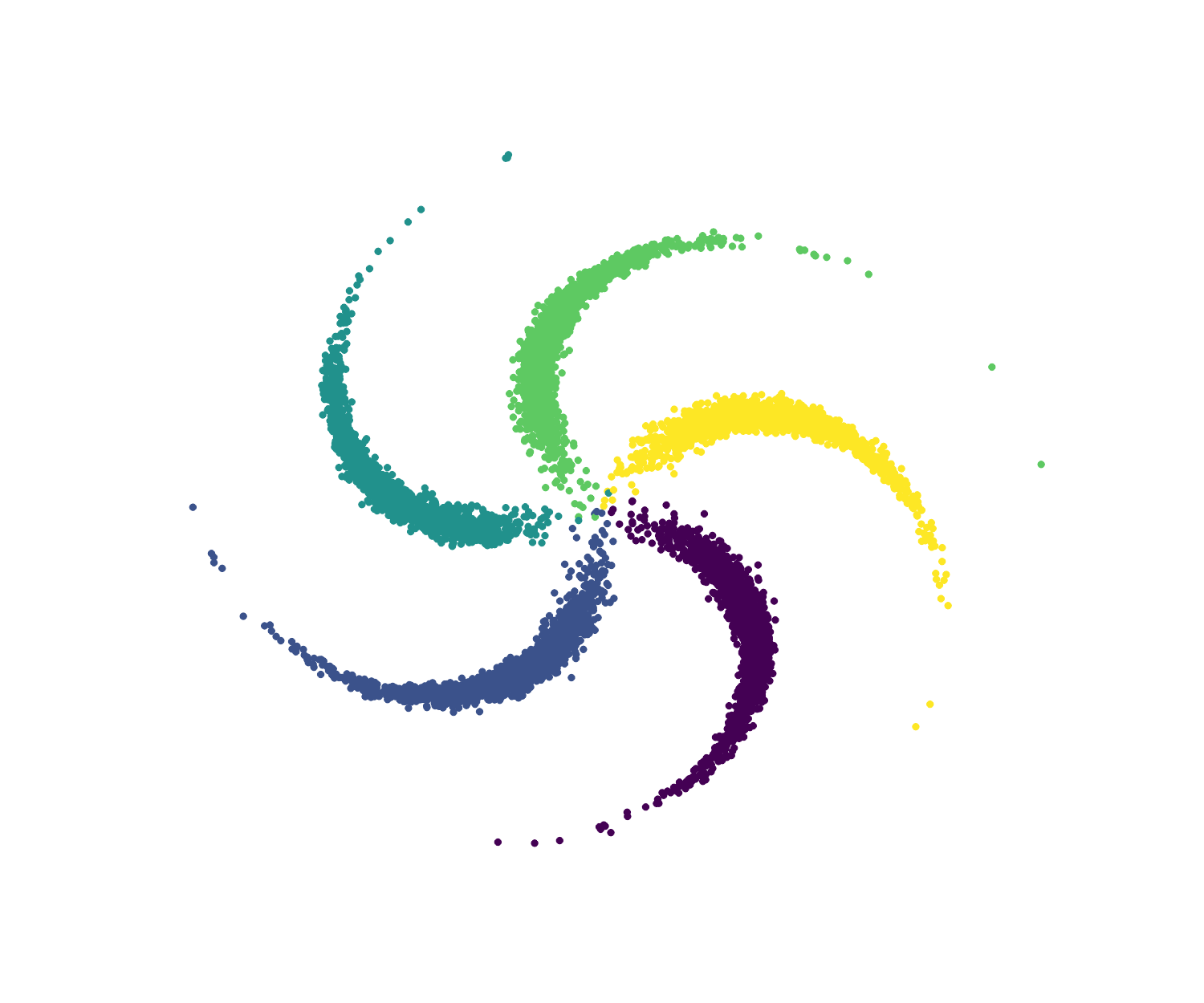}
         \caption{}
         \label{fig: pinwheel_data}
     \end{subfigure}
     \hfill
     \begin{subfigure}[b]{0.32\textwidth}
         \centering
         \includegraphics[width=\textwidth]{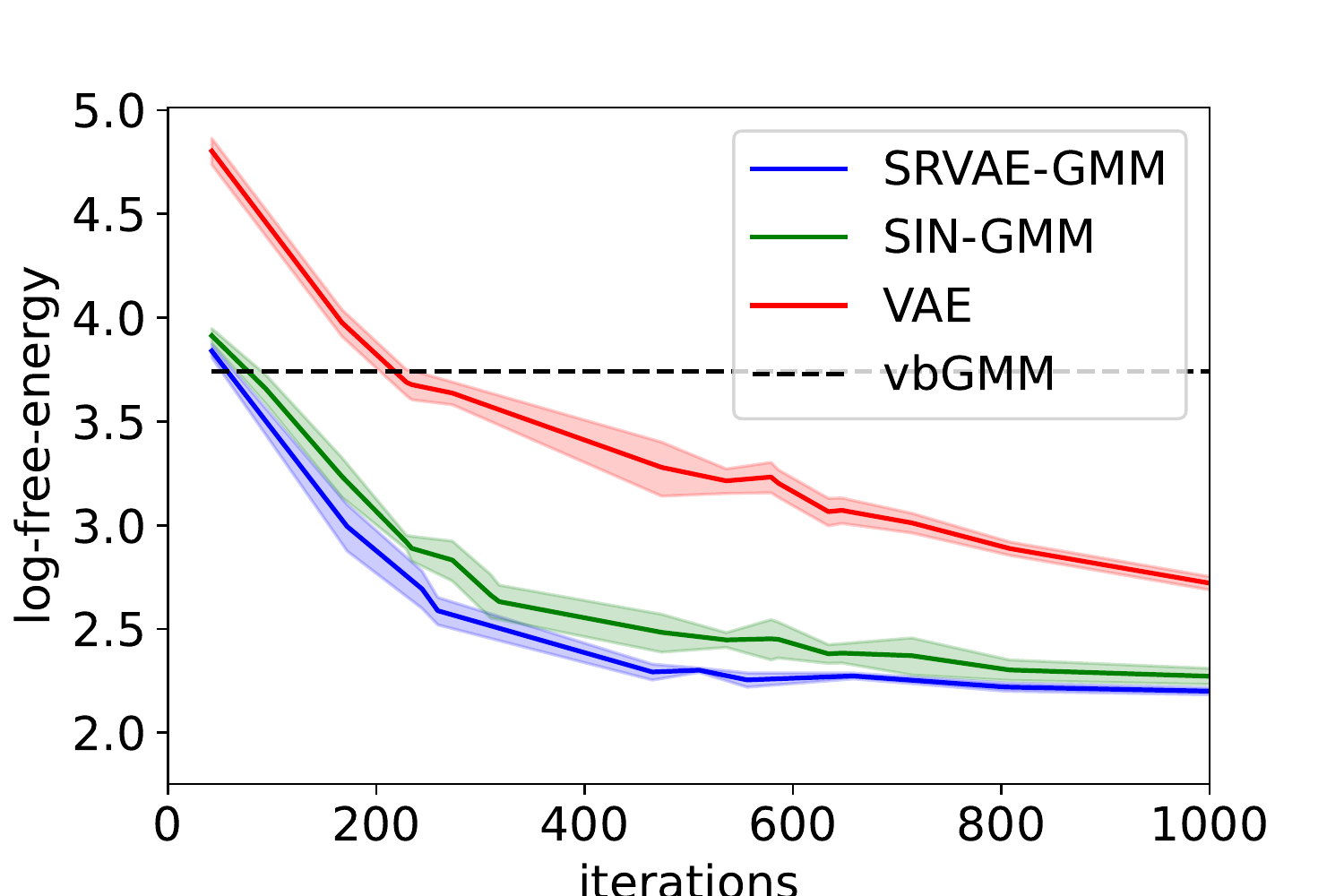}
         \caption{}
         \label{fig: elbo_pinwheel}
     \end{subfigure}
     \hfill
     \begin{subfigure}[b]{0.32\textwidth}
         \centering
         \includegraphics[width=\textwidth]{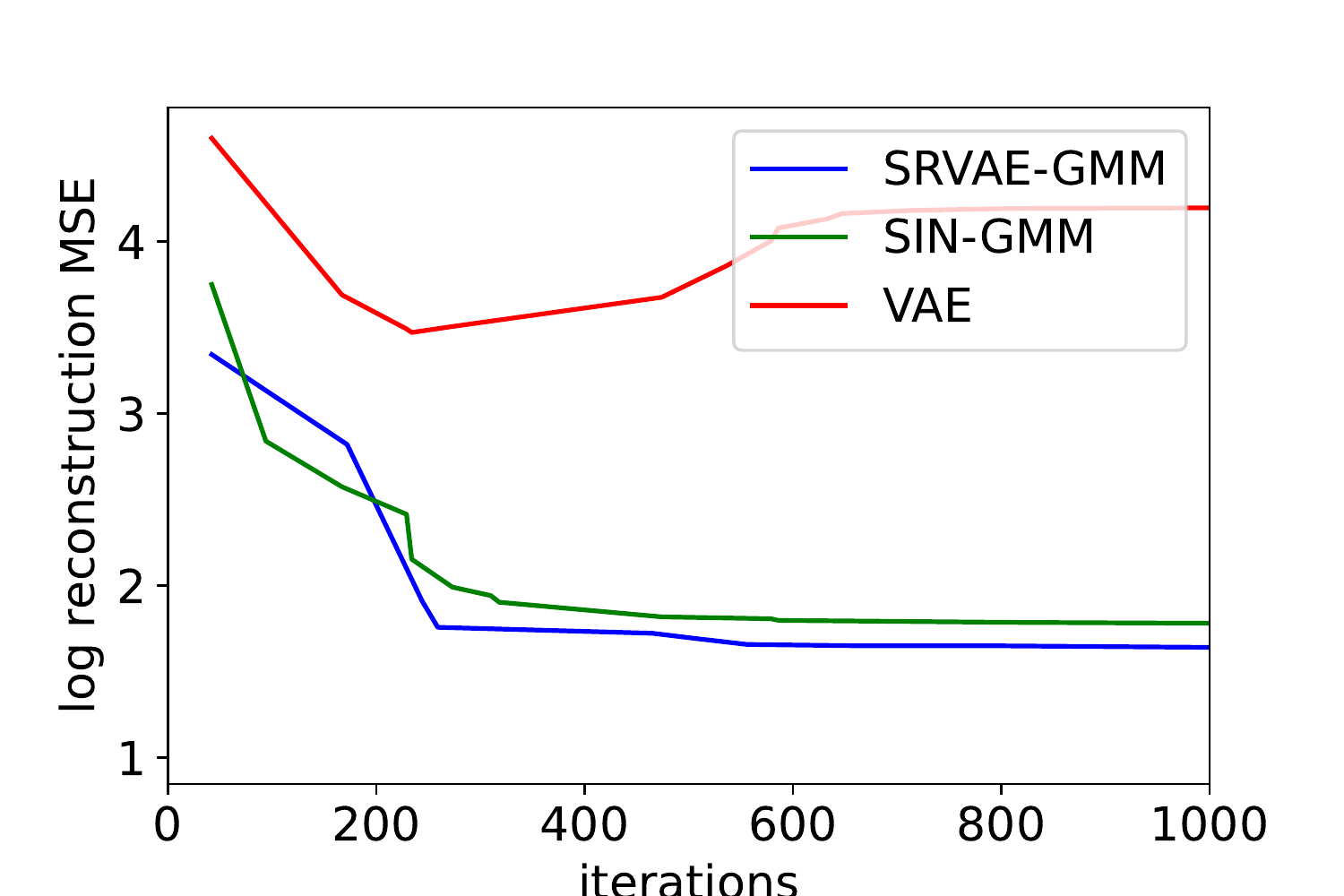}
         \caption{}
         \label{fig: recon_pinwheel}
     \end{subfigure}
     \hfill
     \begin{subfigure}[b]{0.32\textwidth}
         \centering
         \includegraphics[width=\textwidth]{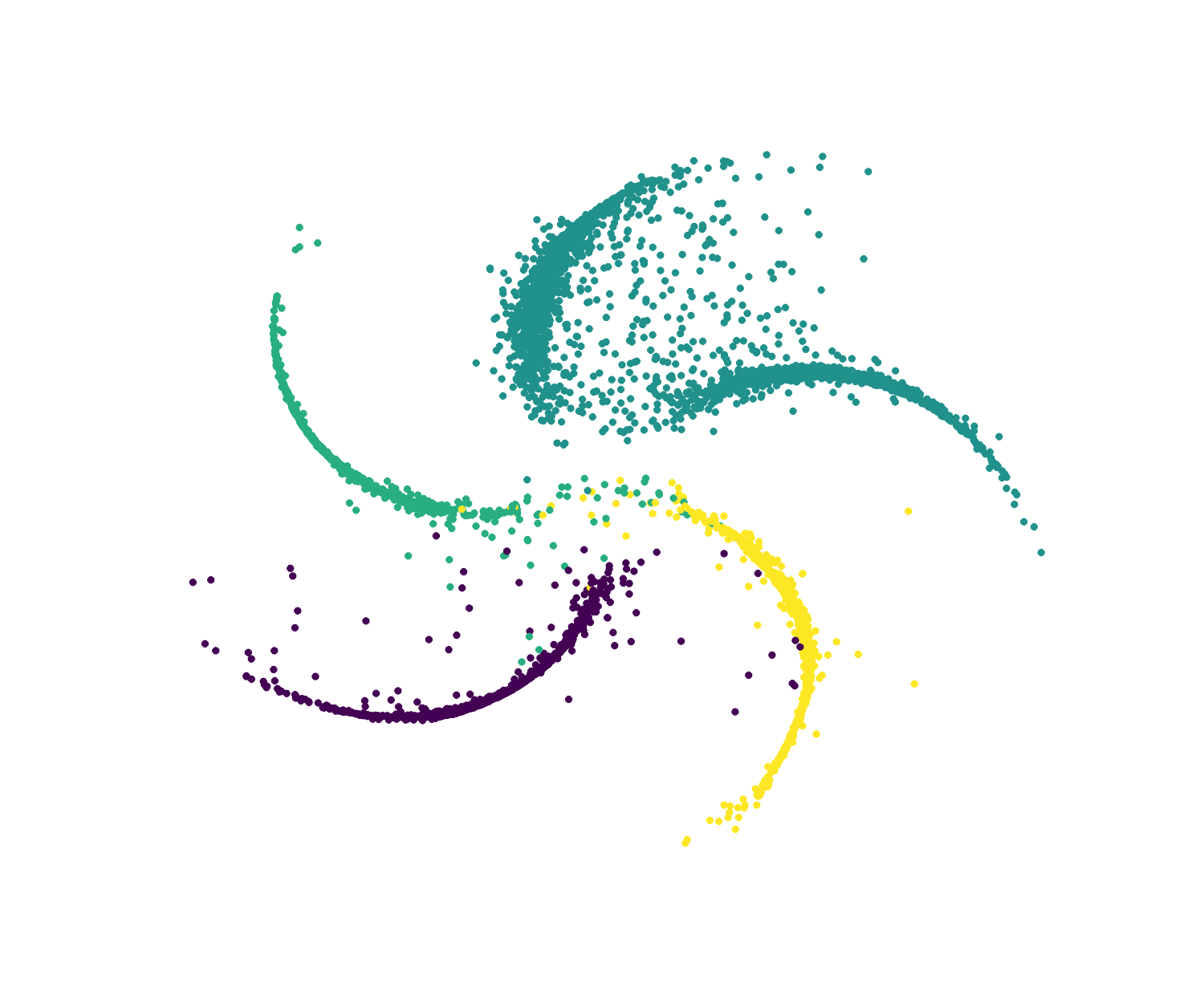}
         \caption{}
         \label{fig: aea_pinwheel_generation}
     \end{subfigure}
     \hfill
     \begin{subfigure}[b]{0.32\textwidth}
         \centering
         \includegraphics[width=\textwidth]{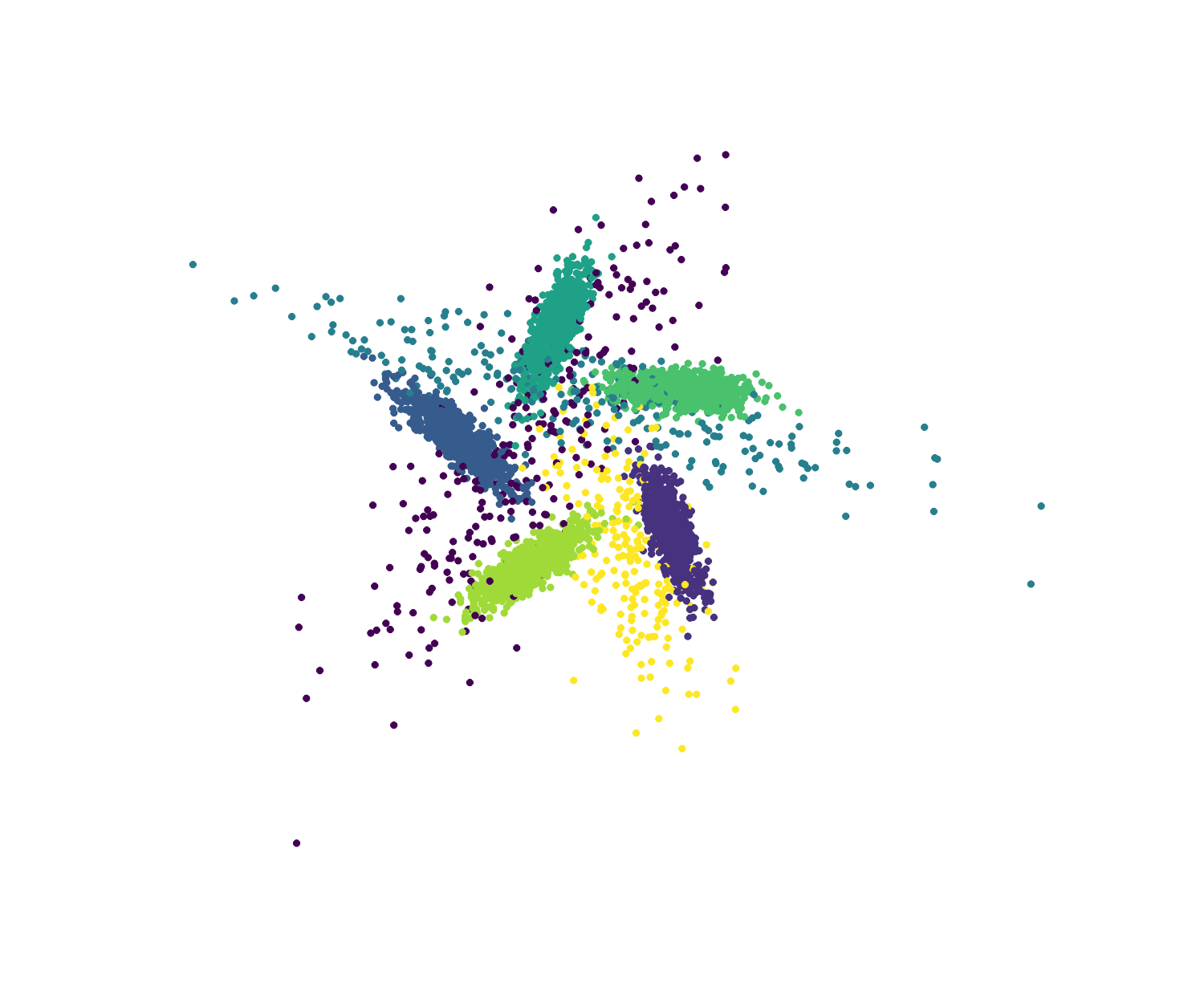}
         \caption{}
         \label{fig: gmm_pinwheel_generation}
     \end{subfigure}
     \hfill
     \begin{subfigure}[b]{0.32\textwidth}
         \centering
         \includegraphics[width=\textwidth]{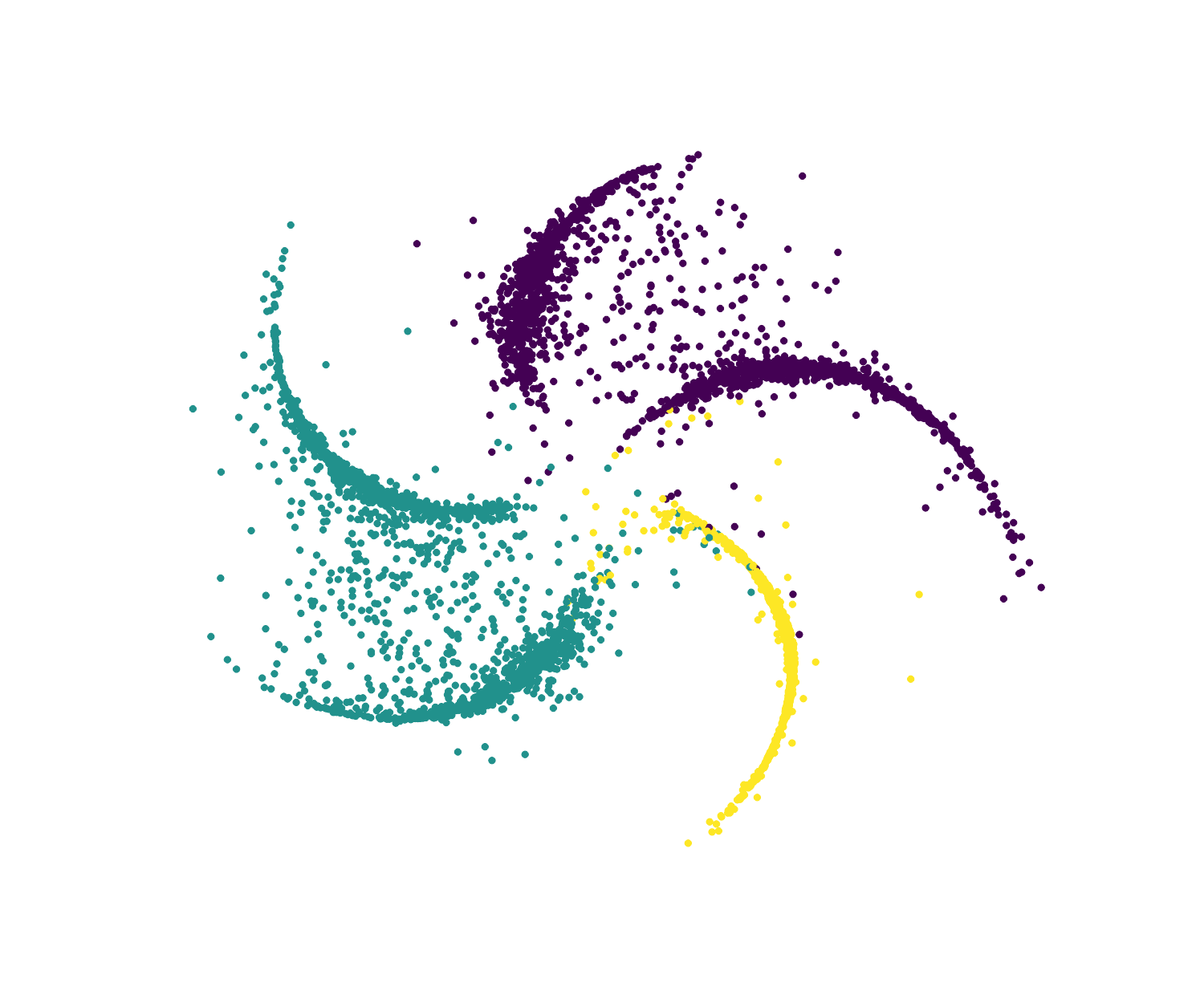}
         \caption{}
         \label{fig: sin_pinwheel_generation}
     \end{subfigure}
        \caption{\textbf{Empirical Evaluations on Pinwheel Dataset.} (a) Sampled data from the pinwheel dataset; Training curves of the (b) variational free energy objective and (c) testing reconstruction MSE through out training (all in log-scale); Sampled data given trained model of (d) SRVAE-GMM; (e) variational GMM; (f) SIN-GMM~\cite{lin2018variational}. All evaluations are based on the averages over 5 random seeds.
        }
        \label{fig: pinwheel_results}
\end{figure}

To assess the empirical performance of structured amortisation in a standard latent-variable model, we evaluated
SRVAE-GMM on the classic pinwheel dataset (Figure~\ref{fig: pinwheel_data})~\cite{johnson2016composing}.
We compare SRVAE-GMM with the structured inference network~\cite[SIN; ][]{lin2018variational}, which
is a more flexible instantiation of the SVAE framework. From Figure~\ref{fig: pinwheel_results}, we
observe that SRVAE-GMM outperforms SIN-GMM in terms of the training variational free energy,
reconstruction mean-squared error, and generation fidelity.
The models are identical apart from the recognition stage, providing support for the idea that by
capturing the ``explaining away'' posterior factors, structured recognition leads to better training
and stronger generation.




\subsection{Experiments with SR-nlGPFA}
\subsubsection{Synthetic and EEG Dataset}
\label{sec: aea_svgpfa_eeg}

We compared SR-nlGPFA to methods that do not capture the inter-latent posterior correlation, using
synthetic data and a small-scale EEG dataset used
previously~\cite{requeima2019gaussian,ashman2020sparse}.  Unless stated otherwise, we employed an
exponentiated quadratic kernel, $\kappa(x, x') = \lambda \exp(-\frac{||x-x'||^2}{\tau^2})$, where
$\lambda$ and $\tau$ are the marginal variance and length-scale parameters, respectively.

\textbf{Baselines} Our main baseline model is SGP-VAE~\cite{ashman2020sparse} (described in
Sections~\ref{sec: aea_svgpfa} and \ref{sec: related_works}).
We also compared to a ``vanilla'' VAE with fully factorised Gaussian variational
distribution~\cite{kingma2013auto, rezende2014stochastic}, and SVAE with a linear dynamical system
(SVAE-LDS) latent prior~\cite{johnson2016composing}.
All models were implemented with the same recognition and generative network architectures (see
Appendix~\ref{sec: app_implementation_detail} for implementation details).

\textbf{Synthetic Dataset} We generated data from the GPFA generative model (Eq.~\ref{eq:
  gpfa_gen}), with $g(\cdot) = \sigma(\Phi(\cdot))$, where $\Phi(\cdot)$ represents the functional
mapping through a fixed, randomly-initialised 2-layer MLP with ReLU hidden non-linearity, and
$\sigma(\cdot)$ is the sigmoid function.

\textbf{EEG Dataset} We follow the experimental procedure described
by~\citet{requeima2019gaussian}, and consider an EEG measurement dataset spanning $1$ second at
256\,Hz sample frequency, taken during image viewing~\cite{zhang1995event}. Each datapoint
consists of the voltage readings from $7$ electrodes positioned on the participant's scalp.  Here we
report results when
all data are observed, contrary to the settings in~\cite{requeima2019gaussian, ashman2020sparse} (we
include results and further discussion on partial observability in Appendix~\ref{sec:
  app_aea_svgpfa}).


{
\begin{table}\small
  \renewcommand{\arraystretch}{1.2}
  \begin{tabular}{cccccc} 
     & \multicolumn{2}{c}{\textbf{Synthetic data}} & \multicolumn{2}{c}{\textbf{EEG}} & {\textbf{Population spiking}}
    \\
    \cmidrule(lr){2-3}\cmidrule(lr){4-5}\cmidrule(lr){6-6}
    & SMSE & NLL & SMSE & NLL & SMSE\\
    SR-nlGPFA & $\mathbf{0.31\pm0.02}$ & $\mathbf{1.69\pm 0.04}$ & $\mathbf{0.27\pm 0.05}$ & $\mathbf{1.81\pm0.21}$ & $\mathbf{0.47\pm0.18}$ \\ 
    SGP-VAE~\cite{ashman2020sparse} & $0.36\pm 0.05$ & $1.76\pm 0.07$ & $0.35\pm 0.05$ & $2.17\pm 0.17$ & $0.55\pm 0.19$ \\ 
    Vanilla-VAE~\cite{kingma2013auto, rezende2014stochastic} &  $1.05\pm 0.02$ & $13.60\pm 3.58$ & $0.57\pm 0.09$ & $3.45\pm 0.87$ & $3.19\pm 1.98$ \\ 
    SVAE-LDS~\cite{johnson2016composing} &  $0.933\pm 0.02$ & $9.37\pm 1.94$ & $3.04\pm 0.38$ & $11.71\pm 2.66$ & $2.52\pm 1.31$ \\ 
  \end{tabular}
  \vspace{1ex}
  \caption{Quantitative comparison of performance between SR-nlGPFA and baseline models on synthetic dataset, EEG dataset and population spiking dataset (Section~\ref{sec: aea_svgpfa_neural}) with respect to standarised mean squared error (SMSE) and negative log-likelihood (NLL).  Averages over $5$ random seeds.}
  \label{tab: synthetic_eeg_results}
\end{table}
}


SR-nlGPFA achieved lower standardised mean squared error (SMSE) and lower test negative
log-likelihood (NLL) than the alternative methods on both data sets (Table~\ref{tab:
  synthetic_eeg_results}), providing evidence for the benefits of structured recognition.  The gains
come despite the computational complexity of SR-nlGPFA being of the same order as SGP-VAE (see
appendix~\ref{sec: app_aea_svgpfa}).

\subsubsection{Population Neuronal Firing Data}
\label{sec: aea_svgpfa_neural}

The firing of place cells in Hippocampal area CA1 and grid cells in the medial Entorhinal Cortex
(mEC) is known to be modulated by the animal's location~\cite{o1978hippocampus, moser2008place},
speed and direction of locomotion~\cite{mcnaughton1983contributions, sargolini2006conjunctive,
  kropff2015speed}, with the mapping between these behavioural covariates and neural activity
expressed in non-linear mixed tuning curves.
The behavioural signals are continuous and often mutually dependent, and so create temporal and
spatial structure in the time series of population activity.
We asked whether SR-nlGPFA and related methods would be able to identify and extract this structure
without supervision; that is, without direct access to the behavioural covariates.

We used single-cell spiking data from neurons in the hippocampal CA1 and mEC regions of rats
recorded during exploration of a Z-shaped track, as reported by~\citet{olafsdottir2016coordinated}.
The data comprised $28$ experimental sessions, each spanning $10$ minutes. Example neural firing
patterns of the population are shown in Figure~\ref{fig: exemplary_pc_fp} of the Appendix.


For SR-nlGPFA and SGP-VAE, we adopted the GPFA generative model Eq.~\ref{eq: gpfa_gen} with DGM non-linearity and Poisson observation
likelihood.
\begin{equation}
    p(\vec{y}(x)|\vec h(x)) = \prod_{n=1}^{N}\text{Poisson}(y_{n}(x) | g(\vec{h}(x))_{n})
\label{eq: poisson_likelihood}
\end{equation}
where $x$ are  discrete times, $N$ is the number of recorded neurons, $g(\cdot)$ represents the neural-network mapping from the GPFA features, $\vec{h}$, to the rate of Poisson-distributed firing counts for each neuron over contiguous 100\,ms bins.
SR-nlGPFA obtained a lower SMSE of prediction than SGP-VAE and other methods tested (Table~\ref{tab: synthetic_eeg_results}).


Computational constraints meant that the SR-nlGPFA model was trained using short batches of data and 64 inducing points per batch.   
Latent trajectory estimates derived from such batches will not necessarily be continuous at the boundaries between them.
Thus, once the model was fit, we performed svGP inference over complete sessions with increased numbers of inducing points (see Section~\ref{sec: aea_svgpfa} and Appendix~\ref{sec: app_aea_neural}).

To relate these recovered latent time-series to behavioural covariates, we performed two-dimensional Canonical Correlation Analysis~\cite[CCA;][]{hardoon2004canonical}. 
We present results for one session here, emphasising qualitative effects. 
Further qualitative and quantitative results across sessions can be found in
Appendix~\ref{sec: app_aea_neural}.

Figure~\ref{fig: cc_corr_pc_1} shows a heatmap of the correlation coefficients between the canonical correlates of the posterior means ($CCX\{1, 2\}$) and  the individual behavioural covariates (distance from one end of the track, speed, direction of travel, head direction, and `unfolded' position along a full lap of the track).
Many correlations are high, indicating the low-dimensional manifold parametrised by the conjunctive set of behavioural correlates can be accurately captured by the posterior latent variables learned with SR-nlGPFA solely from neural spikes. Please refer to Appendix Figure~\ref{fig: cca_heatmap_all} for numerical values of the correlations (and for other sessions).

To see whether the learned latent structure contained decodable information about behavioural covariates, we extracted the direction-modulated neurons predicted by the trained model (details in Appendix~\ref{sec: app_aea_neural}). Figure~\ref{fig: model_predict_corr_dir} compares the direction modulation of the model-predicted neurons against that of the rest of the neurons, where direction modulation is defined as the correlation between raw single-cell spike counts and the direction values. Neurons predicted by the model to be direction modulated exhibited significantly greater direction modulation than the other neurons (p-value=$5.56 \times 10^{-4}$)~\cite{cacucci2004theta}. Figure~\ref{fig: hd_pred_pc_dir} shows the firing profile for travel in each direction for the neuron with the strongest predicted direction modulation. 
The pattern of firing fields exhibits clear directional dependence. Similar comparisons hold for spatial and speed modulation, indicating that SR-nlGPFA is able, in a purely unsupervised fashion, to learn a latent space that contains linearly decodable information about the behavioural covariates associated with individual neuronal firing, even for neurons whose activity exhibits conjunctive coding.

Figure~\ref{fig: cc_pos_aea} shows the two canonical correlates obtained from the latent trajectories as a function of the animal's spatial location.
These reflect both location and direction of movement (colours), indicating the latent dimensions learned by SR-nlGPFA disentangle direction from spatial location in a simple linear projection, despite the conjunctive coding mechanism exhibited in the CA1 neurons. A similar plot derived from SGP-VAE (Figure~\ref{fig: cc_pos_orig}), shows less clear disentanglement.

Figure~\ref{fig: cc_trackDist_aea} shows the temporal evolution of the latent CCs for SR-nlGPFA and SGP-VAE, as well as the spatial location of the rat. Both $CCX1$ and $CCX2$ of SR-nlGPFA consistently track the dynamics of the distance (with perfect phase alignment and half-cycle offset, respectively), whereas neither $CCX1$ or $CCX2$ of SGP-VAE could accurately reflect the distance. Moreover, given the additional svGP inference step over the entire trajectory (with increased number of inducing points), latent signals of SR-nlGPFA exhibit significantly greater smoothness than that of SGP-VAE.

\begin{figure}[t!]
     \centering
    \hspace{-40pt}
     \begin{subfigure}[b]{0.22\textwidth}
         \centering
         \includegraphics[width=1.5\textwidth]{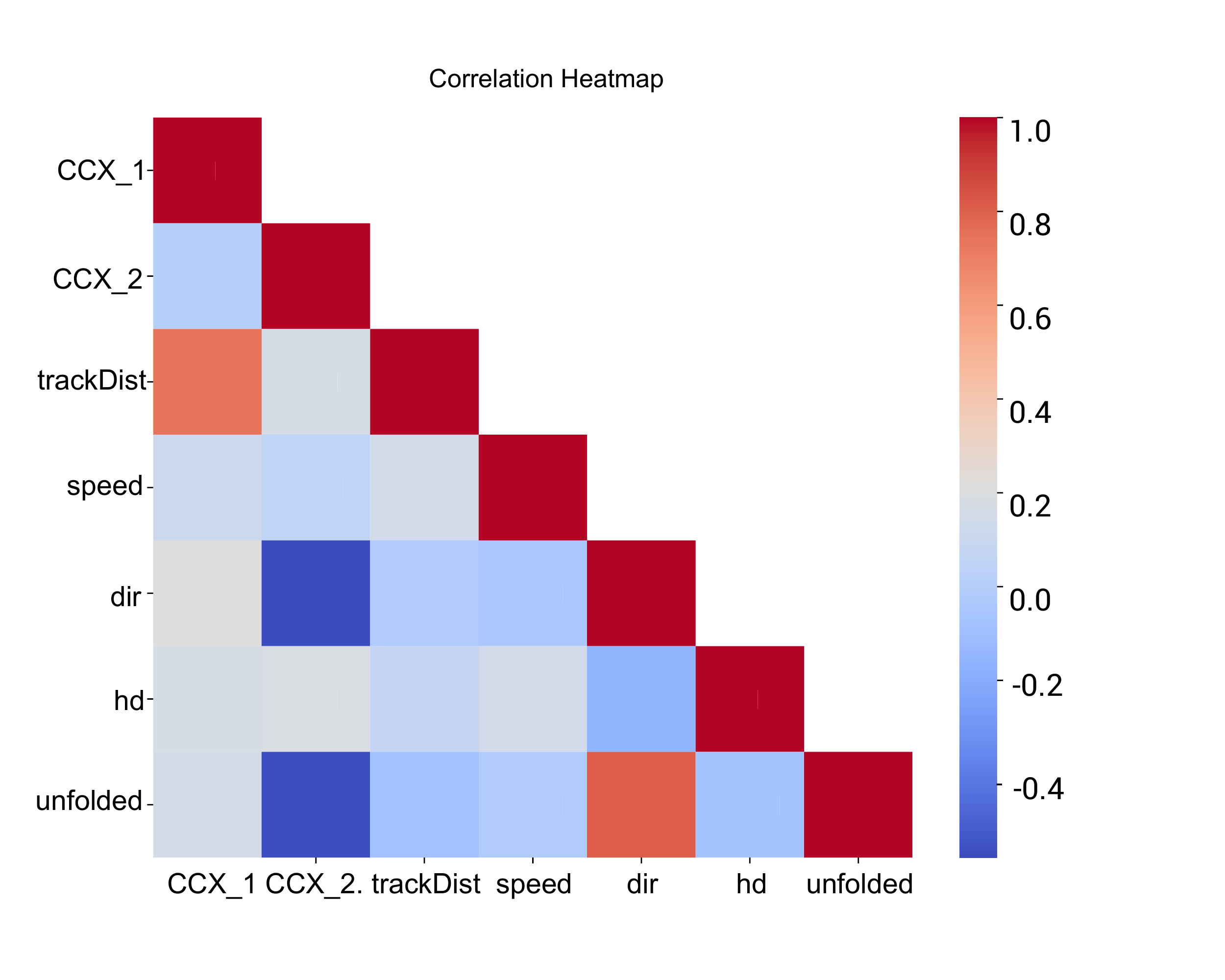}
         \caption{}
         \label{fig: cc_corr_pc_1}
     \end{subfigure}
     \qquad
     \begin{subfigure}[b]{0.26\textwidth}
         \centering
         \includegraphics[width=1.5\textwidth]{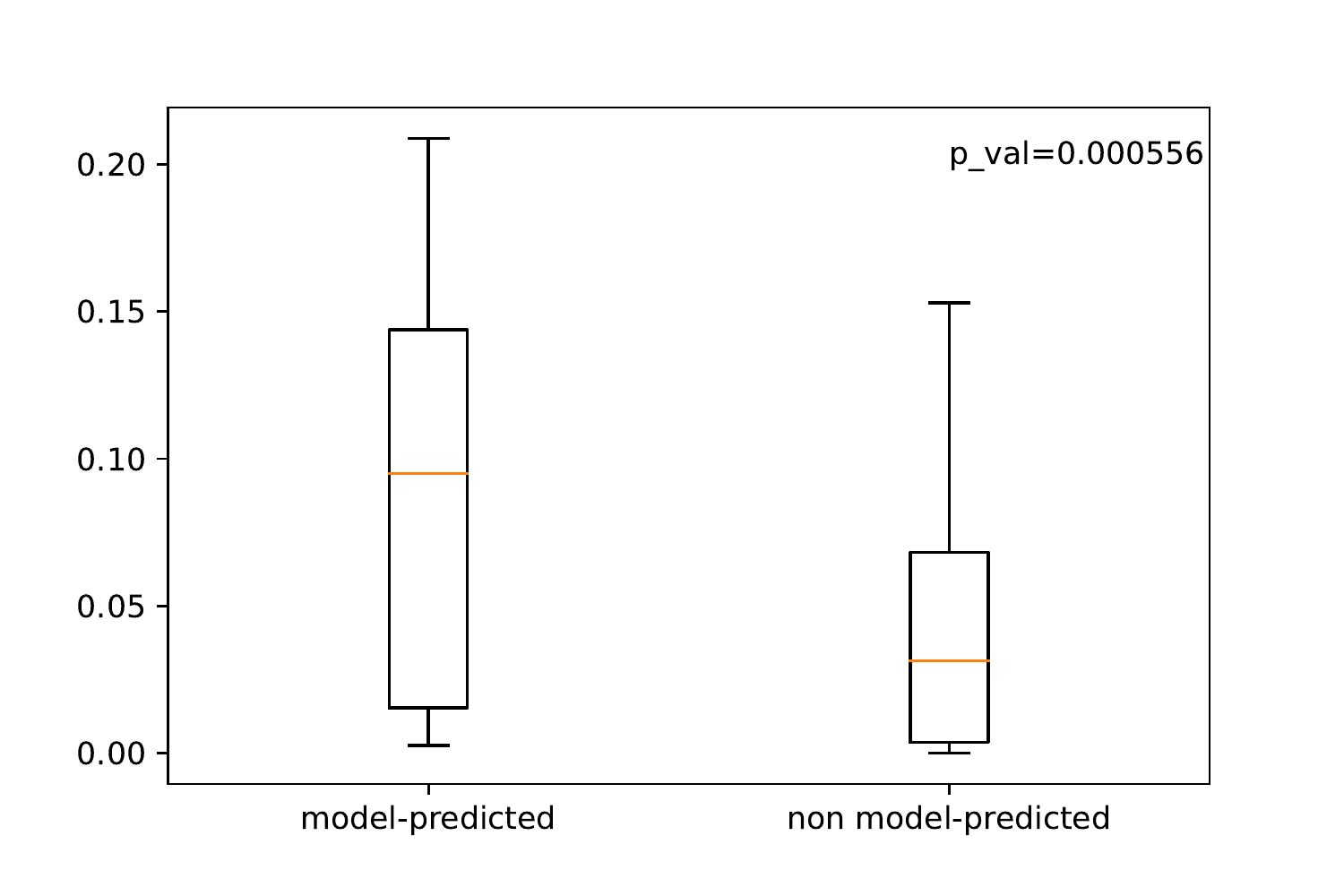}
         \caption{}
         \label{fig: model_predict_corr_dir}
     \end{subfigure}
    \quad 
     \begin{subfigure}[b]{0.48\textwidth}
         \centering
         \includegraphics[width=1.6\textwidth]{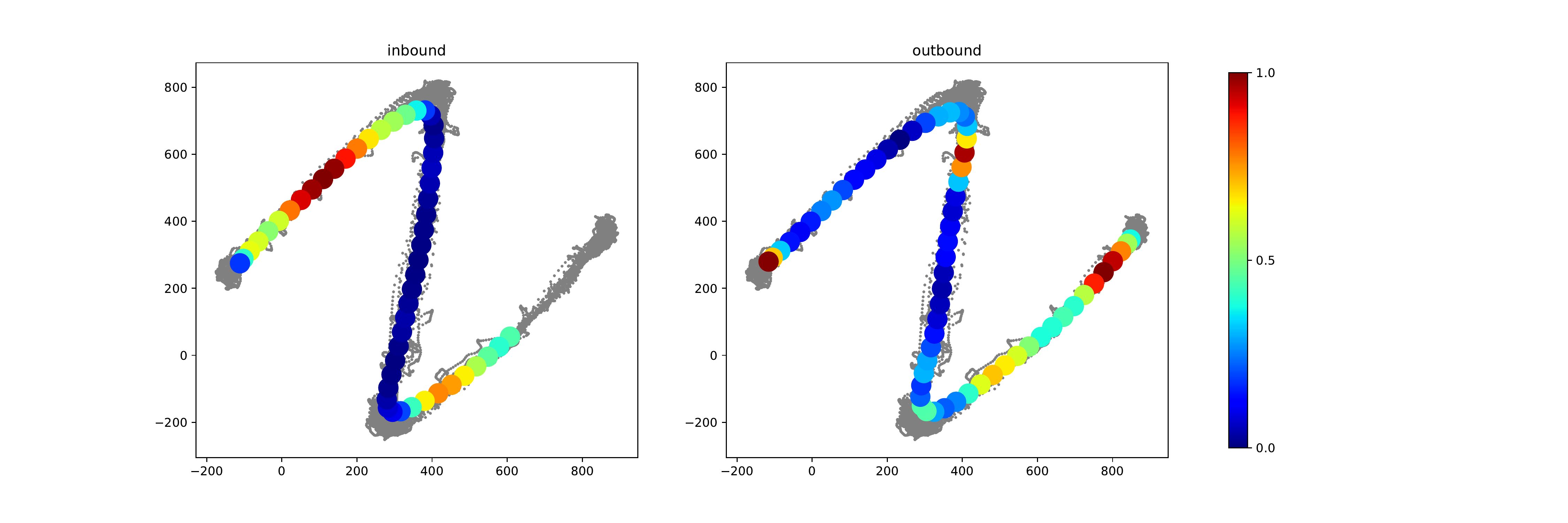}
         \caption{}
         \label{fig: hd_pred_pc_dir}
     \end{subfigure}
     \begin{subfigure}[b]{0.54\textwidth}
         \centering
         \includegraphics[width=1.2\textwidth]{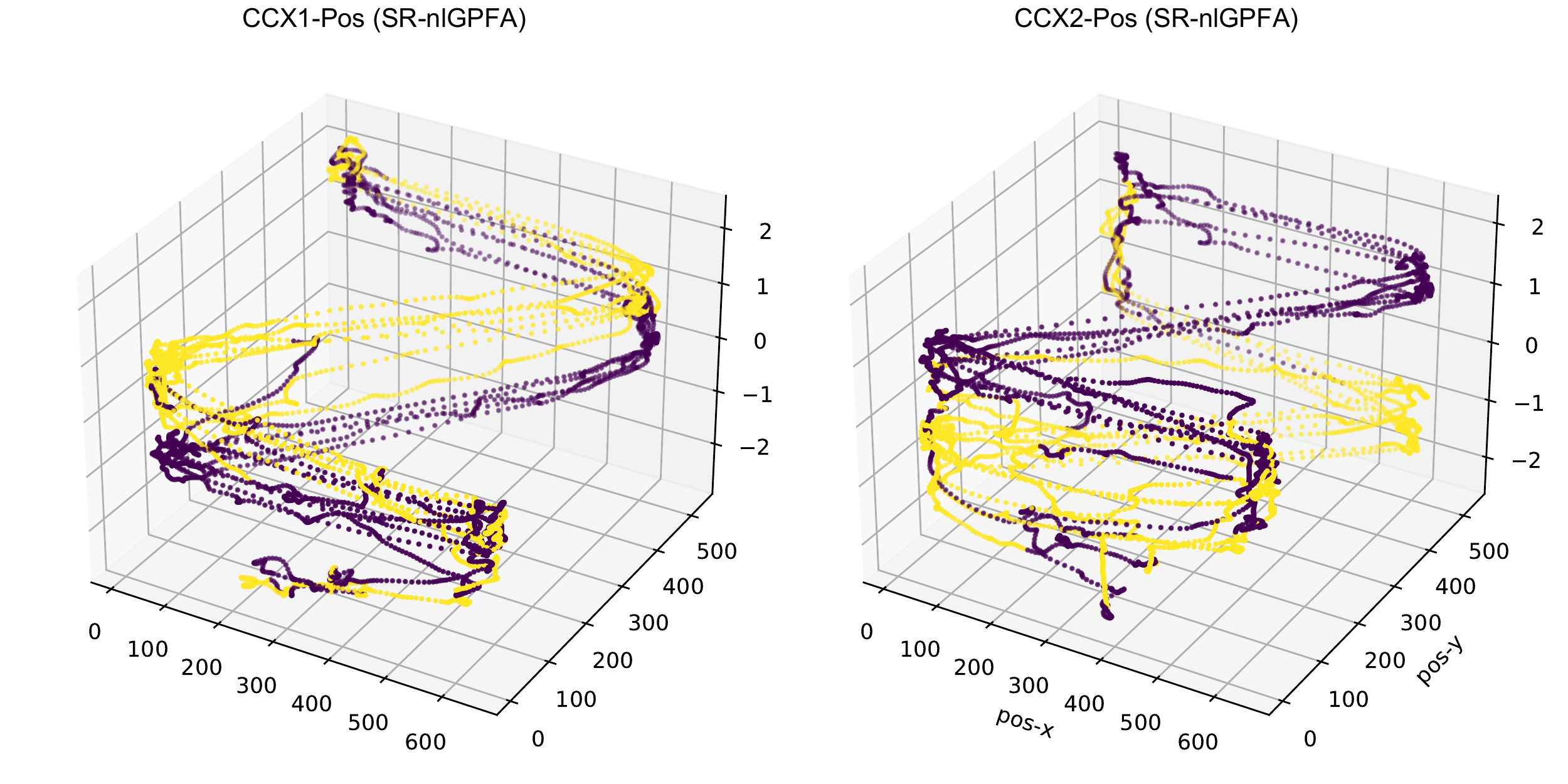}
         \caption{}
         \label{fig: cc_pos_aea}
     \end{subfigure}
     \hfill
     \begin{subfigure}[b]{.35\textwidth}
     \begin{subfigure}[b]{\textwidth}
         \centering
         \includegraphics[width=1.\textwidth]{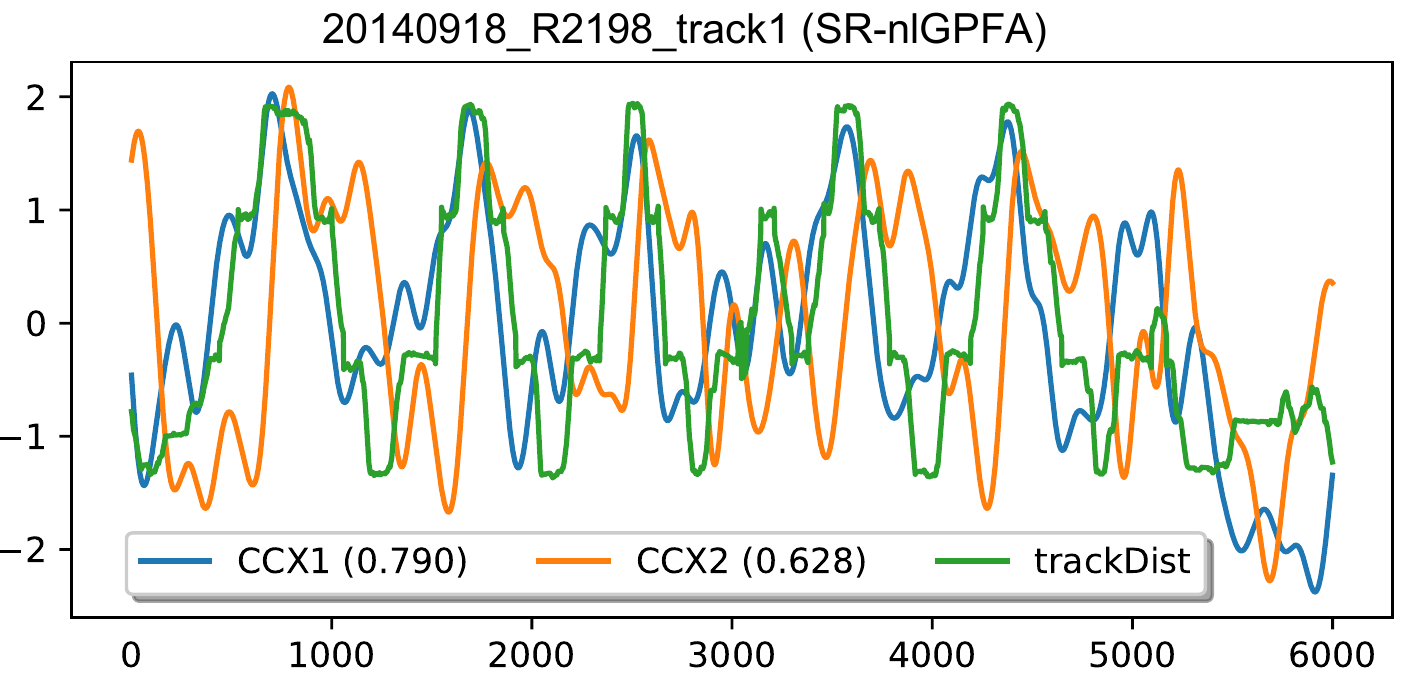}
         \caption{}
         \label{fig: cc_trackDist_aea}
     \end{subfigure}
     \hfill
     \begin{subfigure}[b]{\textwidth}
         \centering
         \includegraphics[width=1.\textwidth]{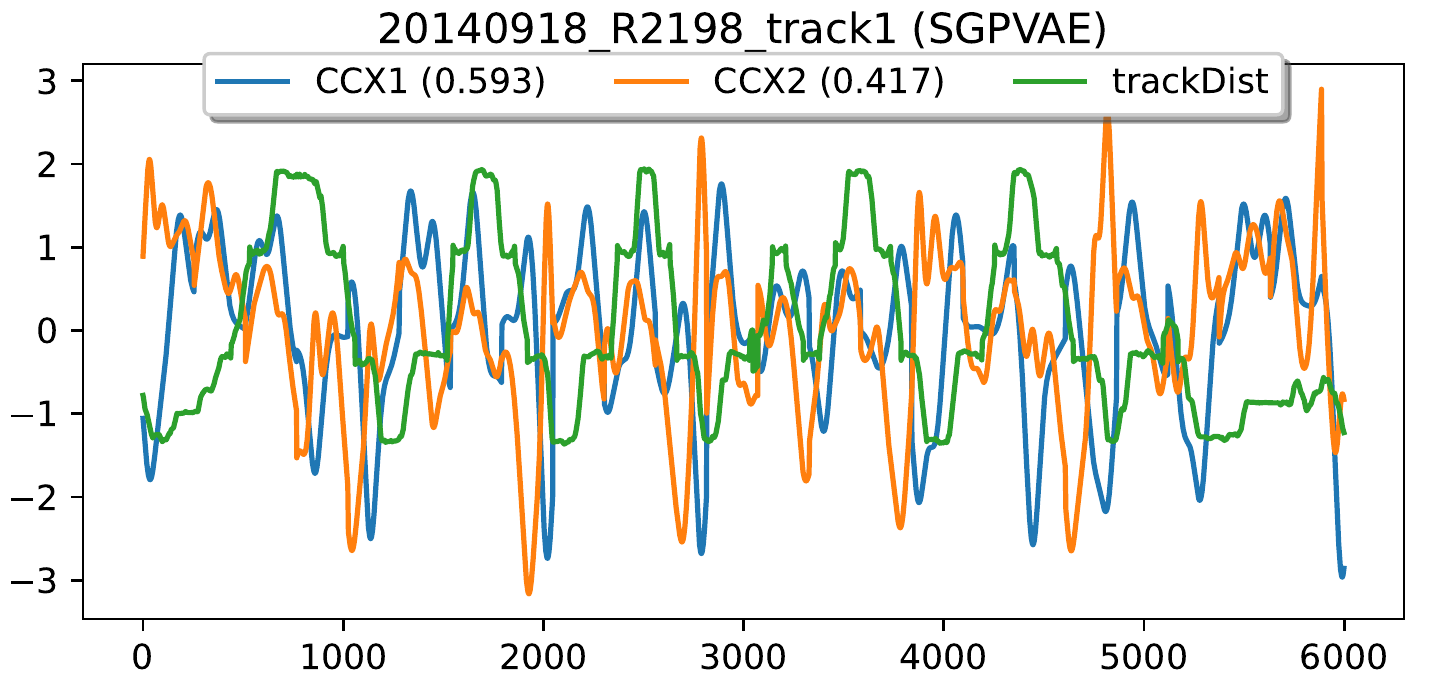}
         \caption{}
         \label{fig: cc_trackDist_orig}
     \end{subfigure}
     \end{subfigure}
     
        \caption{\textbf{Analysis of SR-nlGPFA posterior latent structure}. \textbf{(a)} Heatmap of the correlation between the CCs and the behavioural covariates for the selected session; \textbf{(b)} Directional-modulation comparison between the model-predicted direction-modulated neurons 
        against other neurons; \textbf{(c)} Exemplary firing fields of the model-predicted direction-modulated neuron during inbound v.s. outbound movements; \textbf{(d)} Plot of $CCX\{1, 2\}$ of SR-nlGPFA posterior latents against spatial location, color indicates direction of movement (yellow: inbound, magenta: outbound); \textbf{(e-f)} Temporal evolution of (standardised values of) $CCX\{1, 2\}$ and spatial location for SR-nlGPFA and SGP-VAE.
        }
        \label{fig: spike_data_results}
\end{figure}

\section{Discussion}
We have developed a framework of structured recognition for variational autoencoders, which can be viewed as a strict generalisation of the SVAE~\cite{johnson2016composing}, capturing posterior correlations induced by the ``explaining away" effect. Correspondingly, the SRVAE framework yields a tighter bound on the log-likelihood than the SVAE free energy objective.

Applying structured recognition and non-linear variational autoencoding to the GPFA framework, we introduce SR-nlGPFA, facilitating scalable free-form variational Gaussian approximation. 
SR-nlGPFA supports a fully non-linear, non-conjugate extension to standard GPFA~\cite{yu2008gaussian,adam2016scalable,duncker2018temporal}, which is particularly  well-suited to settings where posterior inference must be scaled to additional data.
We show that SR-nlGPFA outperforms the baseline on all presented tasks, both quantitatively and qualitatively. Studying population spiking data of hippocampal neurons, we show that SR-nlGPFA is able to identify latent signals that exhibit strong correlation with the behavioural covariates that are well-known to be conjunctively encoded by the recorded neurons, but without direct access to these covariates. While here we focused on temporal correlation exhibited in time-series data, we note that the sparse amortised variational approximation presented here can also be straightforwardly applied to other GP-latent models, such as GP-LVM~\cite{lawrence2003gaussian, wu2017gaussian}. While we have considered spiking data alone, these could be substituted by or combined with EEG, calcium imaging, or indeed behavioural data recorded during the same session. Such joint analysis of latent structures given different recordings within the same session could potentially facilitate new findings with respect to, e.g., replay, phase coding, etc.

\section{Acknowledgement}
We thank the anonymous reviewers for  helpful comments and discussions, and
Zilong Ji and Dan Bush for helpful discussions regarding the experiments with
the neural data. This work is funded by the UKRI, DeepMind, the Gatsby
Charitable Foundation, the Simons Foundation and the Wellcome Trust.


\newpage
\bibliography{example_paper}
\bibliographystyle{unsrtnat}

\newpage 
\section*{Checklist}


\begin{enumerate}

\item For all authors...
\begin{enumerate}
  \item Do the main claims made in the abstract and introduction accurately reflect the paper's contributions and scope?
    \answerYes
  \item Did you describe the limitations of your work?
    \answerYes
  \item Did you discuss any potential negative societal impacts of your work?
    \answerNA
  \item Have you read the ethics review guidelines and ensured that your paper conforms to them?
    \answerYes
\end{enumerate}

\item If you are including theoretical results...
\begin{enumerate}
  \item Did you state the full set of assumptions of all theoretical results?
    \answerYes
        \item Did you include complete proofs of all theoretical results?
    \answerYes
\end{enumerate}

\item If you ran experiments...
\begin{enumerate}
  \item Did you include the code, data, and instructions needed to reproduce the main experimental results (either in the supplemental material or as a URL)?
    \answerYes
  \item Did you specify all the training details (e.g., data splits, hyperparameters, how they were chosen)?
    \answerYes
        \item Did you report error bars (e.g., with respect to the random seed after running experiments multiple times)?
    \answerYes
        \item Did you include the total amount of compute and the type of resources used (e.g., type of GPUs, internal cluster, or cloud provider)?
    \answerYes
\end{enumerate}

\item If you are using existing assets (e.g., code, data, models) or curating/releasing new assets...
\begin{enumerate}
  \item If your work uses existing assets, did you cite the creators?
    \answerYes
  \item Did you mention the license of the assets?
    \answerYes
  \item Did you include any new assets either in the supplemental material or as a URL?
    \answerNA
  \item Did you discuss whether and how consent was obtained from people whose data you're using/curating?
    \answerNA
  \item Did you discuss whether the data you are using/curating contains personally identifiable information or offensive content?
    \answerNA
\end{enumerate}

\item If you used crowdsourcing or conducted research with human subjects...
\begin{enumerate}
  \item Did you include the full text of instructions given to participants and screenshots, if applicable?
    \answerNA
  \item Did you describe any potential participant risks, with links to Institutional Review Board (IRB) approvals, if applicable?
    \answerNA
  \item Did you include the estimated hourly wage paid to participants and the total amount spent on participant compensation?
    \answerNA
\end{enumerate}

\end{enumerate}
\newpage 
\appendix

\section{Details on Structured VAE}
\label{sec: app_svae_further}

The variational free energy for VAE takes a similar formulation as the non-amortised version:
\begin{equation}
    \mathcal{F}(\theta, \phi) = \angles[\big]{\log p(y, z| \Theta)-\log q(z|y, \phi)}_{q(z|y, \phi)}
\end{equation}
where we utilise a variational distribution parametrised by a recognition network, $q(z|y, \phi)$. The parametric assumptions in standard VAE formulation limits the expressiveness of the latent space modelling, hence leading sub-optimal training and variational inference. In order to improve the flexibility of both the generative modelling and the variational approximation, \citet{johnson2016composing} proposed structured VAE, combining PGM-parametrised latent prior distribution with amortised inference such that the flexibility of neural network modelling and the conditional dependence structure can be integrated to derive the variational distribution.

Specifically, the structured autoencoding framework considers the following free energy:
\begin{equation}
    \mathcal{F}(\Theta, \lambda) = \angles[\Big]{\log \frac{p(y| z, \gamma)p(z| \theta_{\text{PGM}})}{q(z|\lambda)}}_{q(z|\lambda)}
\end{equation}
It is clear to see that the partial optimisation on $q(z|\lambda)$ yields the following (partially) optimal variational approximation:
\begin{equation}
    q(z|\lambda^{\ast}) \propto p(y|z, \gamma)p(z|\theta_{\text{PGM}})
\end{equation}
Due to the apparent intractability, a recognition network is used to generate amortised approximation of the generative likelihood:
\begin{equation}
    q(z|\lambda^{\ast}) \propto p(y|z, \gamma)p(z|\theta_{\text{PGM}}) \approx r(z|y, \phi)p(z|\theta_{\text{PGM}})
\end{equation}
where $r(z|y, \phi)$ is the recognition potential parametrised by $\phi$. With conjugate mean-field recognition potential, the resulting $q(z|\lambda^{\ast})$ can hence be computed analytically via VMP, and thus contains the factored structure inherent in the prior PGM, whilst exhibits flexibility due to the neural network parametrised recognition potential. The generative and recognition parameters can be trained following standard stochastic optimisation by maximising the \textit{surrogate} variational free energy:
\begin{equation}
    \mathcal{F}_{\text{SVAE}}(\Theta, \phi) = \angles[\Big]{\log \frac{p(y| z, \gamma)p(z| \theta_{\text{PGM}})}{q(z|\lambda^{\ast})}}_{q(z|\lambda^{\ast})}
\end{equation}
Note that in the original SVAE formulation, \citet{johnson2016composing} consider hyperpriors on $\theta_{\text{PGM}}$, and are updated given natural gradient descent. Here we assume deterministic $\theta_{\text{PGM}}$ for simplicity (except for the SRVAE-GMM model), and the structured amortisation framework can be easily extended to adapt to the variational Bayes setting.

\section{More Instantiations of Structured Recognition Framework with Latent Variable Models}
\label{sec: app_other_models}

Here we provide two additional instantiations of the SRVAE framework. The corresponding empirical evaluations can be found in Appendix~\ref{sec: app_further_results}.

\subsection{Tree-Structured Latent PGM}
\label{sec: tree-structured}
\begin{figure}
     \centering
     \begin{subfigure}[b]{0.42\textwidth}
         \centering
         \includegraphics[width=\textwidth]{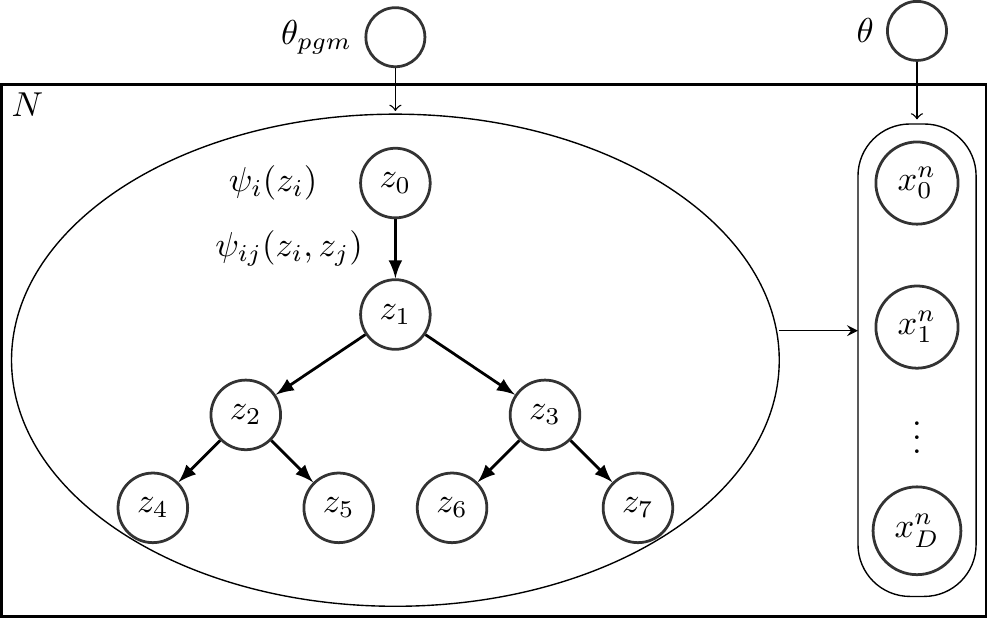}
         \caption{}
         \label{fig: tree_generative}
     \end{subfigure}
     \hfill
     \begin{subfigure}[b]{0.46\textwidth}
         \centering
         \includegraphics[width=\textwidth]{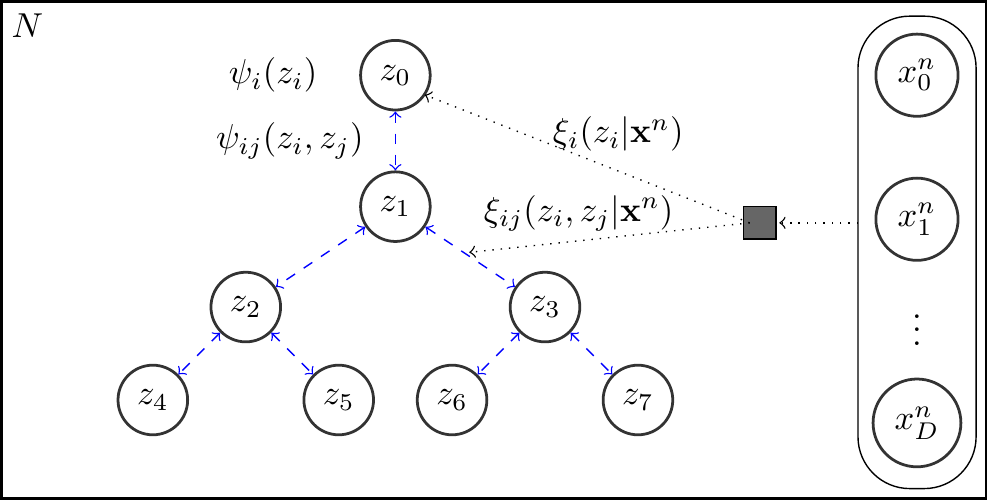}
         \caption{}
         \label{fig: tree_inference}
     \end{subfigure}
        \caption{\textbf{SRVAE framework implemented with tree-structured latent variable model.} \textbf{(a)} generative model \textbf{(b)} inference model, with tree-structured (or a joint factor over all latent variables) recognition potentials.}
        \label{fig: tree_pgm}
\end{figure}

We instantiate the SRVAE framework with discrete tree-structured PGM
(Figure~\ref{fig: tree_generative}), which we term as TreeSRVAE.
A general tree-structured PGM takes the following density function.
\begin{equation}
    p(z) = \frac{1}{Z}\prod_{i}\psi_{i}(z_{i})\prod_{(i, j)\in E}\psi_{ij}(z_{i}, z_{j}) 
    \label{eq: tree_prior}
\end{equation}
Non-linear generative likelihood functions introduces joint factors that does not exist a prior, known as ``explaining away" (Section~\ref{sec: method}). Below we provide two potential solutions

The optimal variational distribution should have the same PGM structure as the posterior, which often contains joint factors over many (possibly all) latent variables.
However, as the number of states grows exponentially with the number of latent variables, in practice it is usually not scalable to have amortised inference output a joint factor that approximates the exact posterior joint factor potentials. We hence seek scalable alternatives. Here we design the recognition network to output the amortised factor potential to be of the same structure as the prior PGM; in this case, a tree-structured distribution with the same set of nodes and pairwise connections (Figure~\ref{fig: tree_inference}).
\begin{equation*}
    r(z|y; \phi) = \prod_{i}\xi_{i}(z_{i})\prod_{(i, j)\in E}\xi_{ij}(z_{i}, z_{j}) 
\end{equation*}

Note that we do not require the recognition network to output a ``proper" density function, but only the factored potentials up to some normalising constant (singleton and pairwise in TreeSRVAE). By utilising a amortised potential of the same structure of the prior distribution, we could easily perform the partial optimisation step of combining the recognition output with the prior distribution, by performing updates only on the existing factored structures.
\begin{equation}
\begin{split}
    &q^{\ast}(z|x, \theta, \phi)\propto\prod_{i}\psi'_{i}(z_{i})\prod_{i, j}\psi'_{ij}(z_{i}, z_{j}) \\ 
    &\text{ where } \psi'_{i}(z_{i}) = \psi_{i}(z_{i})\xi_{i}(z_{i}), \forall i, \quad \text{ and } \psi'_{i, j}(z_{i}, z_{j}) = \psi_{i, j}(z_{i}, z_{j})\xi_{i, j}(z_{i}, z_{j}), \forall (i, j)
\end{split}
\end{equation}
The training procedure follows the general VAE formulation, i.e., we need to generate reparametrised samples from the variational distribution to compute the Monte Carlo estimate of the free energy objective.  Thus far we only have the density function up to the normalising constant. Given the tree-structured posterior latent variables, the normalising constant can be computed exactly with Belief Propogation (BP;~\cite{wainwright2008graphical}). 
At each iteration $k$, the message propagation between variable $i$ and factor $a$ are given by
\begin{equation}
    \begin{split}
        \mu^{(k)}_{i\rightarrow a}(z_{i}) &= \prod_{c\in\mathcal{N}(i)\setminus a}\mu_{c\rightarrow i}^{(k-1)}(z_{i})\\
        \mu^{(k)}_{a\rightarrow i}(z_{i}) &= \sum_{\vec{x}_{a}\setminus x_{i}}\psi_{a}(z_{a})\prod_{j\in\mathcal{N}(a)\setminus i}\mu_{j\rightarrow a}^{(k-1)}(z_{j})
    \end{split}
\label{eq: msg_update}
\end{equation}
Due to the tree structure (i.e., no loops), belief propagation converges after a single inward-outward pass (similar to forward-backward propagation in HMM models). Upon convergence, we can compute the marginal distributions of the variables and factors as products of messages and factor potentials.
\begin{equation}
    \begin{split}
        b_{i}(z_{i}) &\equaltoc \prod_{c\in\mathcal{N}(i)}\mu_{c\rightarrow i}^{\ast}(z_{i})\\
        b_{a}(z_{a}) &\equaltoc \psi_{a}(z_{a})\prod_{i\in\mathcal{N}(a)}\mu_{i\rightarrow a}^{\ast}(z_{i})
    \end{split}
\end{equation}

The singleton and pairwise marginal beliefs can be used to sample from the relevant components of $q^{\ast}(z|x, \theta, \phi)$. Due to the tree-structure, the order of sampling the latent variables is irrelevant. That is, we can initiate ancestral sampling from an arbitrary node $i$ in the tree, computing the conditional distribution $q^{\ast}(z_{j}|z_{i})$ from the relevant marginals. To sample categorical latent variables we employ the Gumbel-Softmax relaxation~\cite{maddison2016concrete, jang2016categorical} so that gradients of the expection can be back-propagated through the samples. 



The free energy then takes the following form.
\begin{equation*}
    \mathcal{F}(\phi, \theta, \gamma) = \mathbb{E}_{q^{\ast}(z|x, \theta, \phi)}[\log p(y|z, \gamma)] - \text{KL}[q^{\ast}(z|x, \theta, \phi)||p(z, \theta)]
\end{equation*}
Given the tree structure, we can compute the KL-divergence analytically with the singleton and pairwise beliefs computed with BP.
\begin{equation}
    \begin{split}
        \text{KL}[q(\vec{z})||p(\vec{z})] &= \sum_{\vec{z}}q(\vec{z})\log{q(\vec{z})} - \sum_{\vec{z}}q(\vec{z})\log{p(\vec{z})}\\
        &= \sum_{\vec{z}}q(\vec{z})\log\left(\frac{\prod_{(i, j)}q_{ij}(z_{i}, z_{j})}{\prod_{i}q_{i}(z_{i})^{(d_{i}-1)}}\right) - \sum_{\vec{z}}q(\vec{z})\log\left(\frac{\prod_{(i, j)}p_{ij}(z_{i}, z_{j})}{\prod_{i}p_{i}(z_{i})^{(d_{i}-1)}}\right)\\
        &= \sum_{i, j}\sum_{\vec{z}}q(\vec{z})\log{q(z_{i}, z_{j})} - \sum_{i}(d_{i}-1)\sum_{\vec{z}}q(\vec{z})\log{q(z_{i})} \\
        & \quad -\sum_{i, j}\sum_{\vec{z}}q(\vec{z})\log{p(z_{i}, z_{j})} + \sum_{i}(d_{i}-1)\sum_{\vec{z}}q(\vec{z})\log{p(z_{i})}\\
        &= \sum_{i, j}\sum_{z_{i}, z_{j}}q(z_{i}, z_{j})\log{q(z_{i}, z_{j})} - \sum_{i}(d_{i}-1)\sum_{z_{i}}q(z_{i})\log{q(z_{i})} \\
        & \quad -\sum_{i, j}\sum_{z_{i}, z_{j}}q(z_{i}, z_{j})\log{p(z_{i}, z_{j})} + \sum_{i}(d_{i}-1)\sum_{z_{i}}q(z_{i})\log{p(z_{i})}\\
        &= \sum_{i, j}\text{KL}[q(z_{i}, z_{j})||p(z_{i}, z_{j})] - \sum_{i}(d_{i}-1)\text{KL}[q(z_{i})||p(z_{i})]
    \end{split}
    \label{eq: tree_kl}
\end{equation}

We note that it is also possible to use other dependency structures of the amortised factor potentials, such as a joint factor potential over all latent variables, or to employ latent structure discovery techniques to infer the latent dependency structure as well as the variational parameters~\cite{attias1999inferring}.


\subsubsection{Gaussian Factors in Tree-Structured Latent PGM}
\label{sec: app_gaussian_factors}
We note that treeSRVAE can also be implemented with continuous latent variables, and here we provide a simple example with Gaussian-distributed latent variables for illustration (note that the joint distribution is also Gaussian). 
When working with continuous latent variables, summation is replaced with integration in the message passing steps (Eq.~\ref{eq: msg_update}). 
Since Gaussian distributions are fully parametrised by the natural parameters (same holds for other exponential family distributions), 
we are able to perform variational message passing by only propagating the messages of the sufficient statistics~\cite{winn2005variational}. Assume the latent prior distribution specified by a Gaussian MRF has a density function of the following format.
\begin{equation}
\begin{split}
    p(\vec{z}) &= \frac{1}{Z}\exp\left(-\frac{1}{2}\vec{z}^T \vec{A}\vec{z} + \vec{h}^T\vec{z}\right) \\
    &= \frac{1}{Z}\prod_{i}\exp\left(-\frac{1}{2}A_{ii}^2z_{i}^2 + b_{i}z_{i}\right)\prod_{ij}\exp\left(-\frac{1}{2}A_{ij}z_{i}z_{j}\right)\\
    &= \frac{1}{Z}\prod_{i}\psi_{i}(z_{i})\prod_{ij}\psi_{ij}(z_i, z_{j})
\end{split}
\end{equation}
where $\vec{A}$ is the precision matrix and $\vec{h}=\vec{A}\vec{\mu}$, with $\vec{\mu}$ being the mean parameter of the joint Gaussian distribution. Note that $\vec{A}$ is also known as the information matrix, which specifies the dependency structure of the latent distribution.
Assume each message at time $k$ is parametrised by a Gaussian distribution, $m^{(k)}_{i\rightarrow j}(x_{j})\sim\mathcal{N}(x_{j}|\mu_{i\rightarrow j}^{(k)}, (\lambda_{i\rightarrow j}^{(k)})^{-1})$. At each iteration $k$, the message updates are shown as following. 
\begin{equation}
\begin{split}
    m_{i\rightarrow j}^{(k)}(z_{j}) &= \int dz_{i} \psi_{ij}(z_{i}, z_{j})\psi_{i}(z_{i})\prod_{k\in\mathcal{N}(i)\setminus j}m_{k\rightarrow i}^{(k-1)}(z_{i})= \mathcal{N}(z_{j}; \mu_{i\rightarrow j}^{(k)}, \lambda_{i\rightarrow j}^{(k)})\\
    \text{where } \lambda_{i\rightarrow j}^{(k)} &= \frac{A_{ij}^2}{\lambda_{i\backslash j}^{(k)}}; \quad \mu_{i \rightarrow j}^{(k)} = \frac{A_{ij}\mu_{i \backslash j}^{(k)}}{\lambda_{i\rightarrow j}^{(k)}}\\
    \text{and } \lambda_{i\backslash j}^{(k)} &= A_{ii} + \sum_{l\in\text{nbr}(i)}\lambda^{(k)}_{l\rightarrow i}; \quad \mu_{i \backslash j}^{(k)} = \frac{A_{ii}b_{i} + \sum_{l\in\text{nbr}(i)}\lambda^{(k)}_{l\rightarrow i}\mu^{(k)}_{l\rightarrow i}}{\lambda_{i\backslash j}^{(k)}}
\end{split}
\end{equation}
Then the singleton and pairwise beliefs (used for sampling and computing the KL divergence) can be computed as following.
\begin{equation}
\begin{split}
    b(z_{i}) &= \psi_{i}(z_{i})\prod_{j\in\text{nbr}(i)}m_{j\rightarrow i}(z_{i}) = \mathcal{N}(z_{i}|\mu_{i}, \lambda_{i})\\
    \text{where } \lambda_{i} &= A_{ii} + \sum_{j\in\text{nbr}(i)}\lambda_{j\rightarrow i};\quad \mu_{i} = \frac{A_{ii}b_{i} + \sum_{j\in\text{nbr}(i)}\lambda_{j\rightarrow i}\mu_{j\rightarrow i}}{\lambda_{i}}\\
    b(z_{i}, z_{j}) &= \psi_{ij}(z_{i}, z_{j})\psi_{i}(z_{i})\psi_{j}(z_{j})\prod_{u\in\text{nbr}(i)}m_{u\rightarrow i}(x_{i})\prod_{v\in\text{nbr}(j)}m_{v\rightarrow j}(x_{j}) = \mathcal{N}([z_{i}, z_{j}]^T|\vec{m}_{ij}, \Lambda_{ij})\\
    \text{where }\Lambda_{ij} &= \begin{bmatrix} &A_{ii}+\lambda'_{i\backslash j} &\frac{1}{2}A_{ij}, \quad\frac{1}{2}A_{ji} & A_{jj} + \lambda'_{j\backslash i}\end{bmatrix}, \quad 
    \vec{m}_{ij} = \begin{bmatrix} &b_{i}+\lambda'_{i\backslash j}\mu'_{i\backslash j}\\
    &b_{j}+\lambda'_{j\backslash i}\mu'_{j\backslash i}\end{bmatrix}, \\
    \lambda'_{i\backslash j} &= \sum_{u\in\text{nbr}(i)\backslash j}\lambda_{u\rightarrow j} \mu'_{i\backslash j} = \frac{\sum_{u\in\text{nbr}(i)\backslash j}\lambda_{u\rightarrow j}\mu_{u\rightarrow j}}{\lambda'_{i\backslash j}}
    \end{split}
\end{equation}
Given the prior tree structure (i.e., the information matrix, $A$), the recognition network outputs the sufficient statistics (mean and precision matrix) of a Gaussian distribution with the same structure as the tree-based prior distribution, $\mathcal{N}(\vec{z}|\mu', (A')^{-1})$. Note that instead of parametrising a full-rank precision matrix, which requires $\mathcal{O}(D^2)$ outputs ($D$ is the number of latent variables), by constraining the dependency structure as the prior distribution, the recognition network only need to output $3D-1$ scalars (for each batch), where $2D-1$ contribute towards the non-zero entries in the precision matrix, and $D$ contribute towards the linear term. Such type of parametrisation is similar to the modelling of off-diagonal elements in the covariance matrix with low-rank decomposition, but we note that the off-diagonal entries in the covariance matrix do not convey information about the dependency structure of the latent variables, hence our framework provides stronger interpretability. \citet{dorta2018training} proposes a direct low-rank parametrisation of the precision matrix in a VAE setting, which again lacks interpretation in terms of the latent dependency structure. Moreover, we note that they require the matrix inversion to convert the precision matrix into the covariance matrix for inference and sampling, which requires $\mathcal{O}(D^3)$ complexity, whereas we apply the BP, which on tree-structured PGMs only require $\mathcal{O}(D)$ iterations to converge.



\subsection{Gaussian Mixture Model Latent PGM}
\label{sec: app_aea_gmm}
\begin{figure}
     \centering
     \begin{subfigure}[b]{0.3\textwidth}
         \centering
         \includegraphics[width=\textwidth]{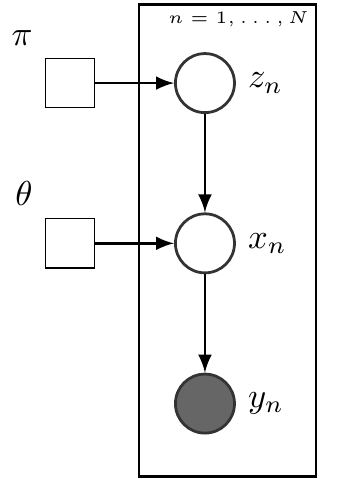}
         \caption{}
         \label{fig: gmm_generative}
     \end{subfigure}
    \qquad
     \begin{subfigure}[b]{0.3\textwidth}
         \centering
         \includegraphics[width=\textwidth]{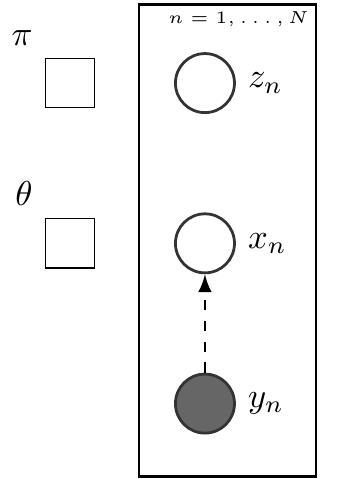}
         \caption{}
         \label{fig: gmm_inference}
     \end{subfigure}
        \caption{\textbf{SRVAEframework implemented with tree-structured latent variable model.} \textbf{(a)} generative model \textbf{(b)} inference model, with tree-structured (or a joint factor over all latent variables) recognition potentials.}
        \label{fig: gmm_pgm}
\end{figure}

The PGMs for the generative and inference models of SRVAE-GMM is shown in Figure~\ref{fig: gmm_pgm}.

\section{Joint Factor Potentials Induced by Non-Linear Likelihood (``Explaining Away")}
\label{sec: app_tree_joint_factor}
Here we show that even a simple non-linear likelihood function can induce non-trivial posterior correlations that do not exist \textit{a priori}.   
Consider a Bernoulli generative likelihood function, with the logits being an affine transformation of $z$.
\begin{equation}
\begin{split}
    p(y|z, \theta) = \mathcal{B}(y; \sigma(Wz + b)) = \prod_{i}\mathcal{B}(y_{i}; \sigma(\sum_{j}W_{ij}z_{j} + b_{i}))
\end{split}
\label{eq: tree_conditional}
\end{equation}
where $\theta = \{W, b\}$. 

Given this simple generative model, and the tree-structured latent distribution (Eq.~\ref{eq: tree_prior}), we can derive the true posterior distribution analytically.
\begin{equation}
    \begin{split}
        \log{p(z|x)} &\equaltoc \log{p(z)}+\log{p(y|z, \theta)} \\ 
         &=\sum_{i}\log{\psi_{i}(z_{i})} + \sum_{i, j}\log{\psi_{i, j}(z_{i}, z_{j})} +\\
         & \qquad \sum_{j}y_{j}\log\left(\frac{1}{1+e^{-\sum_{i}W_{ji}z_{i} + b_{j}}}\right)+(1-y_{j})\log\left(\frac{e^{-\sum_{i}W_{ji}z_{i} + b_{j}}}{1+e^{-\sum_{i}W_{ji}z_{i} + b_{j}}}\right)\\
         &= \sum_{i}\log{\psi_{i}(z_{i})} + \sum_{i, j}\log{\psi_{i, j}(z_{i}, z_{j})} +\\
         & \qquad \sum_{j}y_{j}\left(\sum_{i}W_{ji}z_{i} + b_{j}\right) +  \sum_{j}\log\left(\frac{e^{-\sum_{i}W_{ji}z_{i}+b_{j}}}{1 + e^{-\sum_{i}W_{ji}z_{i}+b_{j}}}\right) \\
         &\equaltoc \sum_{i}\log\phi_{i}(z_{i}) + \sum_{i, j}\log\phi_{i, j}(z_{i}, z_{j}) +\phi_{\vec{z}}(\vec{z})\\
         \text{where } &\phi_{i}(z_{i}) = \psi_{i}(z_{i})\exp(z_{i}\sum_{j}W_{ji}y_{j}), \phi_{ij}(z_{i}, z_{j}) = \psi_{ij}(z_{i}, z_{j}), \forall i, j,\\
         &\quad \phi_{\vec{z}}(\vec{z}) = \sum_{j}\log\left(\frac{e^{-\sum_{i}W_{ji}z_{i}+b_{j}}}{1 + e^{-\sum_{i}W_{ji}z_{i}+b_{j}}}\right)
    \end{split}
    \label{eq: post_mf_deriv}
\end{equation}
We observe that the true posterior distribution involves the singleton and pairwise potentials that preserve the structure of the prior distribution, and at the same time the normalising constant of the conditional likelihood function introduces a joint factor over all latent variables, $\phi_{\textbf{z}}(\vec{z})$, which cannot be captured by the fully factorised amortised potential, but can be partially captured by the tree-structured potentials.

\section{Further Details of SR-nlGPFA}

The derivation of the variational free energy objective for svGPFA (and SR-nlGPFA; Eq.~\ref{eq: svgpfa_free_energy}) is shown as following~\citet{titsias2009variational}.
\begin{equation}
    \begin{split}
        &\log p(\vec{y})\\ 
        &= \log\mltintdx[2]{d\vec{f}\,d\vec{u}_{1:K}} p(\vec{y}, \vec{f}(\cdot), \vec{u}_{1:K})\\
        &= \log\mltintdx[2]{d\vec{f}\,d\vec{u}_{1:K}} p(\vec{y}|\vec{f}(\cdot))\prod_{k}p(f_{k}(\cdot)|\vec{u}_{k})p(\vec{u}_{k}|\vec{z}_{k})\\
        &\geq \mltintdx[2]{d\vec{f}\,d\vec{u}_{1:K}} q(\vec{u}_{1:K}, \vec{f}(\cdot))\log\left[\frac{p(\vec{y}|\vec{f}(\cdot))\prod_{k}p(f_{k}(\cdot)|\vec{u}_{k})p(\vec{u}_{k}|\vec{z}_{k})}{q(\vec{u}_{1:K}, \vec{f})}\right]\\
        &= \mltintdx[2]{d\vec{f}\,d\vec{u}_{1:K}} q(\vec{u}_{1:K}, \vec{f}(\cdot))\log\left[\frac{p(\vec{y}|\vec{f}(\cdot))\prod_{k}p(f_{k}(\cdot)|\vec{u}_{k})p(\vec{u}_{k}|\vec{z}_{k})}{\left[\prod_{k}p(f_{k}(\cdot)|\vec{u}_{k})\right]q(\vec{u})}\right]\\
        &= \mathbb{E}_{q(\vec{h}(\cdot))}\left[\log p(\vec{y}|\vec{h}(\cdot))\right] - \intdx[d\vec{u}_{1:K}] q(\vec{u})\log\frac{q(\vec{u})}{\prod_{k}p(\vec{u}_{k}|\vec{z}_{k})} \\
        &= \mathbb{E}_{q(\vec{h}(\cdot))}\left[\log p(\vec{y}|\vec{h}(\cdot))\right] + \mathcal{H}[q] + \sum_{k}\mathbb{E}_{q(\vec{u}_{k})}[\log p(\vec{u}_{k}|\vec{z}_{k})]
    \end{split}
    \label{eq: gpfa_fe}
\end{equation}
where $\mathcal{H}[q]$ is the entropy of the variational distribution $q(\vec{u})$.

\section{Proof of Proposition 3.1}
\label{sec: app_prop}
We re-state the proposition.
\begin{proposition}
Consider the following latent structured prior.
\begin{equation*}
    p(z; \theta^0) = \frac{1}{Z}\prod_{c\in C}\psi_{c}(z_{c})
\end{equation*}
where we assume $\theta^0$ is the set of trainable deterministic prior parameter. Consider the free energy objective.
\begin{equation*}
    \mathcal{F}[q(z)] = \mathbb{E}_{q(z)}\left[\log\frac{p(z|\theta^0)p(x|z, \theta)}{q(z)}\right]
\end{equation*}
Both the SVAE objective and the AEA-objective take the following expression.
\begin{equation*}
    \mathcal{F}(\theta, \phi) = \mathbb{E}_{q^{\ast}(z)}\left[\log\frac{p(z|\theta^0)p(x|z, \theta)}{q^{\ast}(z)}\right],
\end{equation*}
For SVAE, $q^{\ast}(z)$ is derived through partial optimisation given conjugate amortised inference outputs.
\begin{equation*}
    q^{\ast}(z) = \argmax_{q^(z)} \mathbb{E}_{q(z)}\left[\log\frac{p(z|\theta^0)\prod_{i}l_c(z_{i}|x, \phi)}{q(z)}\right],
\end{equation*}
where $\prod_{i}l_{i}(z_{i}|x, \phi)$ represents the approximate local evidence potentials (corresponding to the product of singleton potentials).

For SRVAE, instead of being fully factorised, the recognition network outputs structured factor potentials, which in principle, could contain factors of arbitrary set of latent variables (e.g., a joint factor for all latent variables).
\begin{equation*}
\begin{split}
    q^{\ast}(z) = \argmax_{q^(z)} \mathbb{E}_{q(z)}\left[\log\frac{p(z|\theta^0)\prod_{c'\in C^{\ast}}r_{c}(z_{c}|x; \phi)}{q(z)}\right],
\end{split}
\end{equation*}
Then the SRVAE objective function provides tighter lower bound to the free energy than the SVAE objective function.
\begin{equation*}
    \max_{q(z)}\mathcal{F}[q(z)] \geq \max_{\phi}\Fsai(\theta, \phi) \geq \max_{\phi} \mathcal{F}_{\text{SVAE}}(\theta, \phi)
\end{equation*}
\end{proposition}

We note that throughout the paper we assume both the SRVAE and SVAE models employ identical generative models, hence we have the stronger results that the inequality for all $\theta$. For SR-nlGPFA, despite introducing the additional affine transformation (parametrised by $\vec C$ and $\vec d$) in the generative process, the linear operation can be completely subsumed into the neural-network generative model, hence leading to the identical generative model as SGP-VAE.

\begin{proof}
The first inequality is trivial. For the second inequality, 
it is easy to see that the set of fully factorised amortised potentials, $L = \{l(z)|l(z)\propto \prod_{i}l_i(z_{i}|x; \phi)\}$, is a strict subset of the set of all structured amortised potentials, $H = \{h(z|x; \phi) | h(z|x; \phi) \propto \prod_{c\in C^{\ast}}r_c(z_{c}|x; \phi) \}$, where $C^{\ast}$ is the set of factors in the amortised approximation to the likelihood. Hence for arbitrary $p(x|z, \theta)$, we have that,
\begin{equation*}
\argmin_{h\in H}\KL{h(z|x, \phi)}{p(x|z, \theta)} \leq \argmin_{l\in L}\KL{\prod_{c\in C}l_c(z_{c}|x, \phi)}{p(x|z, \theta)}
\end{equation*}
with equality if and only if $p(x|z, \theta) \in L$. Hence it is trivial to derive that,
\begin{equation*}
    \max_{\phi}\Fsai(\theta, \phi) = \mathbb{E}_{q^{\ast}_{\text{AEA}}(z)}\left[\log\frac{p(z|\theta^0)p(x|z, \theta)}{q^{\ast}(_{\text{AEA}}z)}\right] \geq \mathbb{E}_{q^{\ast}_{\text{SVAE}}(z)}\left[\log\frac{p(z|\theta^0)p(x|z, \theta)}{q^{\ast}(_{\text{SVAE}}z)}\right] = \max_{\phi}\mathcal{F}_{\text{SVAE}}(\theta, \phi), \forall \theta
\end{equation*}
with equality if and only if $p(x|z, \theta) \in L$. 
\end{proof}

We note that having a tighter free energy lower bound does not necessarily lead to stronger generative modelling~\cite{rainforth2018tighter}. However, by exploring an alternative formulation of the free energy objective (relative to Eq.~\ref{eq: free_energy}): 
\begin{equation}
    \mathcal{F}[q] = \log p(y) - \KL{q(z)}{p(z|y)} \Rightarrow \max_{q}\mathcal{F}[q] = \min_{q}\KL{q(z)}{p(z|y)}
\end{equation}
we see that a tighter free energy bound leads to smaller KL-divergence between the variational approximation and the ground-truth posterior distribution, hence achieving more accurate posterior inference. This is indeed reflected in our empirical evaluation of SR-nlGPFA on the spiking dataset (that the posterior latent processes is significantly more coherent with the underlying behavioural covariates than that of SGP-VAE~\cite{ashman2020sparse}), and also reflected by the better generation quality in the bar dataset (\ref{sec: bar_test}).

\section{Further Experimental Results}
\label{sec: app_further_results}
\subsection{TreeSRVAE on ``Bar" Dataset}
\label{sec: bar_test}
\begin{figure}
     \centering
         \centering
         \includegraphics[width=.3\textwidth]{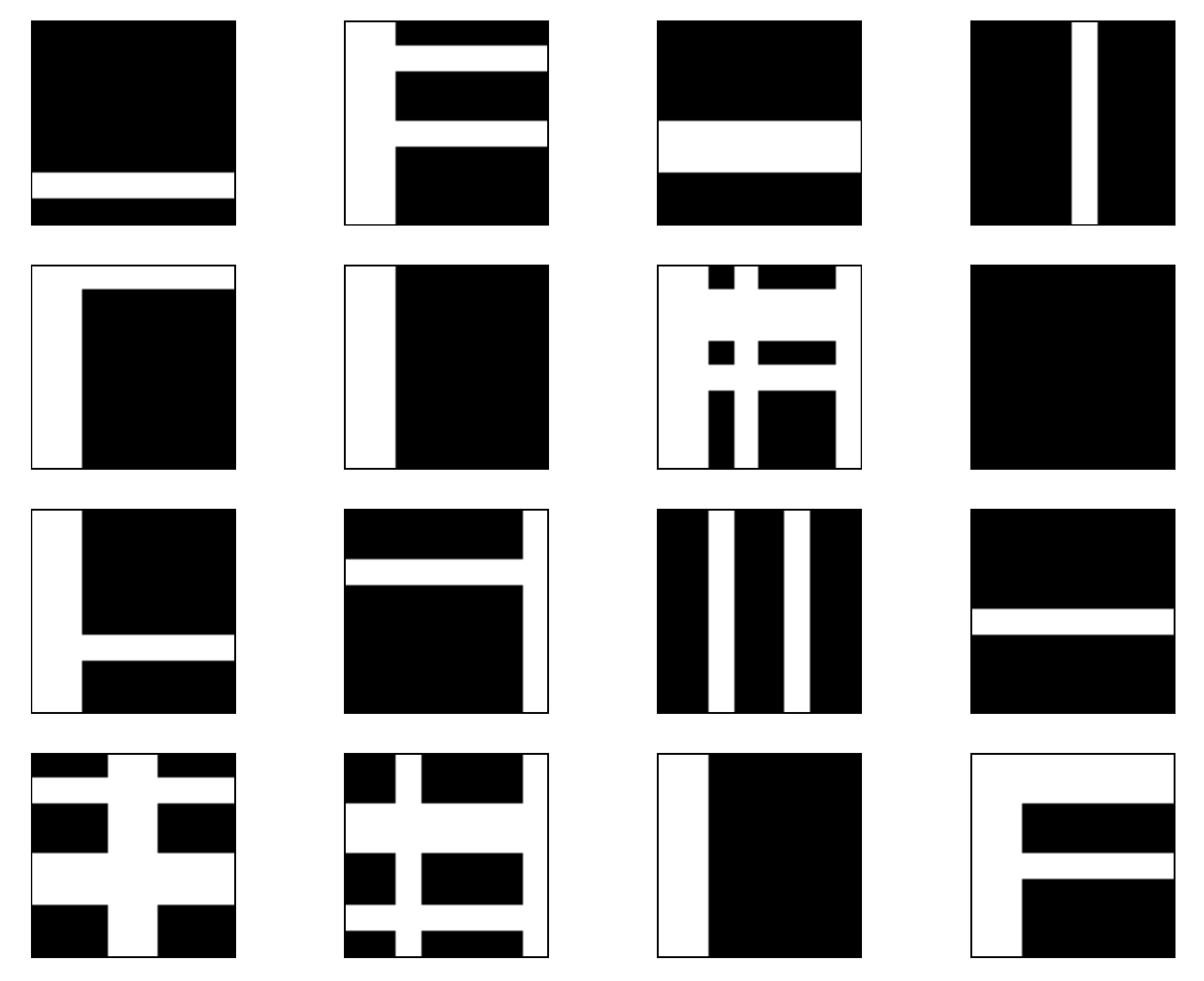}
     
        \caption{\textbf{Samples from the ``Bar" dataset}. 
        }
        \label{fig: bar_test_samples}
\end{figure}

We first evaluate tree-structured recognition potential on a synthetic ``Bar" dataset, consists of square-grid observations with horizontal and vertical bars.
\begin{equation}
\begin{split}
    &p(\vec{z}|\theta) = \mathcal{B}(z_{0}|p_{0})\prod_{i>0}\mathcal{B}(z_{i}|\text{pa}(z_{i})), \\ &p(\vec{y}|\vec{z}, \vec{W}, \vec{b}) = \prod_{d=1}^{D^2}\mathcal{B}(y_{d}|\sigma(\vec{W}\vec{z} + \vec{d})_{d})
\end{split}
\label{eq: bar_generative_model}
\end{equation}
where $\text{pa}(z)$ denote the parent node of node $z$, $\sigma(\cdot)=\frac{1}{1+\exp(\cdot)}$ is the sigmoid function, and $\mathcal{B}(p)$ denotes the Bernoulli distribution with rate $p$. Each binary $z_{i}$ denotes the activation of one bar in the square-grid (e.g., $z_{1} = 1$ indicates the presence of the first horizontal bar). The generative parameters $\vec{W}$ and $\vec{b}$ are defined as 
\begin{equation}
    W_{ij} = \begin{cases}
    2\times \omega_{i} &\text{if pixel i is on the bar j}\\
    0 & \text{otherwise }
    \end{cases}, \quad 
    b_{i} = -\omega_{i}, \text{ for } i = 1, \dots, D^2, 
\end{equation}
where $\{\omega_{i}\}_{i=1}^{D^2}$ are the temperature parameters that control the degree of randomness in the data generation. Note that in current implementations we set $\omega_{i} = \omega$ for all $i$. Exemplary samples from the ``Bar" dataset if shown in Figure~\ref{fig: bar_test_samples} ($D=8)$. 


Note that here we only perform variational inference, instead of variational Bayes optimisation, hence we assume the parameters of the prior distribution (Eq.~\ref{eq: tree_prior}) to be deterministic quantities (which can still be optimised via gradient descent, but do not follow any natural-gradient updates). 


Given the tree-structured prior distribution (Eq.~\ref{eq: bar_generative_model}), we have that the singleton and pairwise potentials take the following expression.
\begin{equation}
\begin{split}
    \psi_{i}(z_{i}) = \begin{cases}
    \mathcal{B}(z_{0}|p^{0}) & \text{if } i = 0;\\
    1 & \text{if } i > 0;
    \end{cases} \quad \psi_{ij}(z_{i}, z_{j}) = \begin{cases}
    \mathcal{B}(z_{i}|p^{i}_{z_{j}}) & \text{if } j = \text{pa}(i);\\
    1 & \text{otherwise };
    \end{cases}
\end{split}
\label{eq: potential_def}
\end{equation}

We use the Gumbel-Softmax trick for relaxed reparametrised sampling of the discrete latent variables.
However, we can also apply a hard-transformation to the relaxed samples to generate the binary-valued latent samples as following.
\begin{equation}
    z_{hard} = 0.5 \times (\text{sign}(z-0.5) + 1)
    \label{eq: hard_sample}
\end{equation}
In practice we find using the hard-sampling trick improves performance for all models considered (VAE, SVAE, TreeSRVAE) on the ``Bar" dataset.

We set the prior distribution to be a "uninformative" prior, where $p^{0} = 0.5$ and $p^{i}_{ij}$ for all $(i, j)$ (Eq.~\ref{eq: potential_def}).
\begin{figure*}[t]
     \centering
     \begin{subfigure}[b]{\textwidth}
         \centering
        \includegraphics[width=1.1\textwidth]{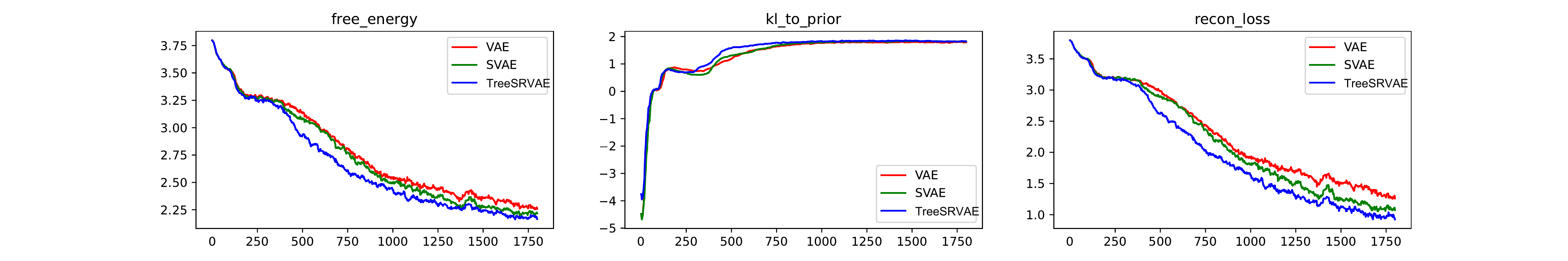}
         \caption{}
         \label{fig: bar_structure_4}
     \end{subfigure}
     \hfill
     \begin{subfigure}[b]{\textwidth}
         \centering
        \includegraphics[width=1.1\textwidth]{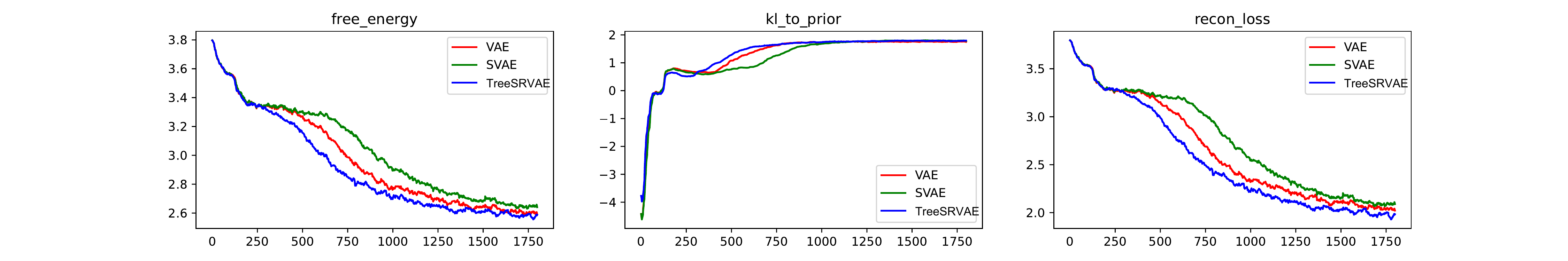}
         \caption{}
         \label{fig: bar_structure_10}
     \end{subfigure}
        \caption{Evaluation of TreeSRVAE on the bar test experiment. \textbf{From Top to Bottom}: Bar-test experiment with (a) sampling temperature of $4$; (b) sampling temperature of $10$. 
        \textbf{From left to right: } free energy; reconstruction loss (log-scale); KL-divergence (with respect to the prior distribution).}
        \label{fig: bar_perf}
\end{figure*}

In Figure~\ref{fig: bar_perf} we show the training curves of the free energy (Eq.~\ref{eq: aea_free_energy}), reconstruction loss ($\log{p(x|z)}$) and KL-divergence (with respect to the prior distribution, $\text{KL}[q^{\ast}(z, \phi)||p(z, \theta_{\text{pgm}})]$) for the ``Bar" dataset generated with varying sampling temperature ($\omega$).
, where the observation is over an $8\times 8$ square grid, hence the latent variables are $z\in\{0, 1\}^{16}$. 
We evaluate on three models: i) the latent variable follows a fully factorised Bernoulli distribution, and the recognition network outputs mean-field inference, which is equivalent to standard VAE with Bernoulli latents~\cite{maddison2016concrete}; ii) the latent variable follows a distribution characterised by a tree-structured probabilistic graphical model, and the recognition network outputs fully factorised potentials, which is equivalent to the original SVAE setup with tree-structured Bernoulli latent~\cite{johnson2016composing}; iii) the latent variable follows a distribution characterised by a tree-structured probabilistic graphical model, and the recognition network outputs tree-structured factor potentials (containing both the singleton and pairwise factor potentials), which is the TreeSRVAE. Note that we use the ``hard-transformation" trick for all models presented in Figure~\ref{fig: bar_perf}.
From Figure~\ref{fig: bar_perf} we see that TreeSRVAE consistently outperforms both VAE and SVAE on the ``Bar" dataset with varying level of sampling noise, in terms of the overall free energy (both the sample efficiency and asymptotic performance), hence providing a tighter lower bound to the true likelihood. 

What if the underlying latent structure deviates largely from being tree-structured? We consider the following ``side-dependent" prior distribution ($p(\vec{y}|\vec{z}, \vec{W}, \vec{b})$ remain the same).
\begin{equation}
    p(\vec{z}|\theta) \propto \text{Cat}(z_{1:D}|p_{1:D})\text{Cat}(z_{D+1:2\times D}|p_{D+1:2\times D})
\label{eq: bar_side_dependent}
\end{equation}
where $\text{Cat}(\vec{p})$ denotes the categorical distribution with probabilities $\vec{p}$. Intuitively, sampling from such prior distribution means that we will observe one and only one bar on each side of the square. Clearly, such prior distribution cannot be modelled by any tree-structured distribution. 

\begin{figure}[t]
         \centering
         \includegraphics[width=.45\textwidth]{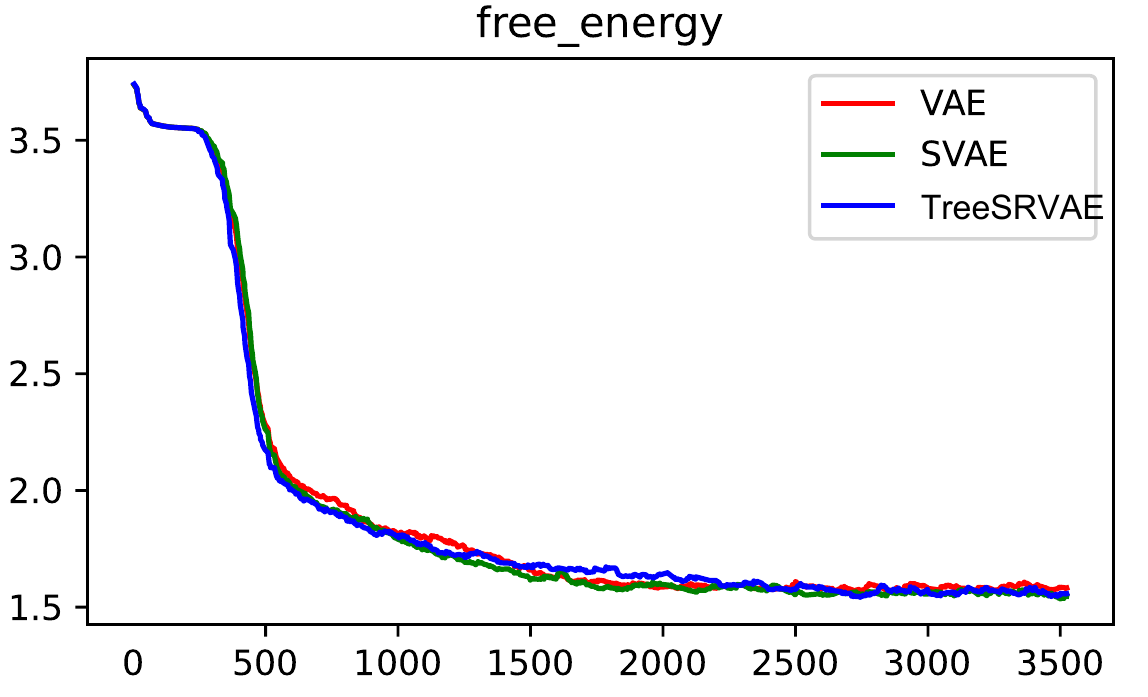}
         \caption{Free energy of the TreeSRVAE, SVAE and VAE over the training process on the ``side-dependent" ``Bar" dataset (Eq.~\ref{eq: bar_side_dependent})}
         \label{fig: bar_uniform_results}
     \end{figure}

We apply the same set of models to the data from the new generative model, and the results are shown in Figure~\ref{fig: bar_uniform_results}. We observe that the performances for all three methods are mostly similar. Hence in situations where there exists a large mismatch between the structures of the amortised potentials and the true posterior distribution, the additional structure present in the amortised inference will not interrupt learning, but instead allow the learning process be at least on the same level of performance as standard VAE and SVAE in terms of free energy. 


Despite having achieved similar free energy through training (Figure~\ref{fig: bar_uniform_results}), we argue that the structured recognition framework should still be preferred. Namely, we examine the generation quality of the generated samples. 

\begin{figure*}[t]
     \centering
     \begin{subfigure}[b]{0.32\textwidth}
         \centering
         \includegraphics[width=\textwidth]{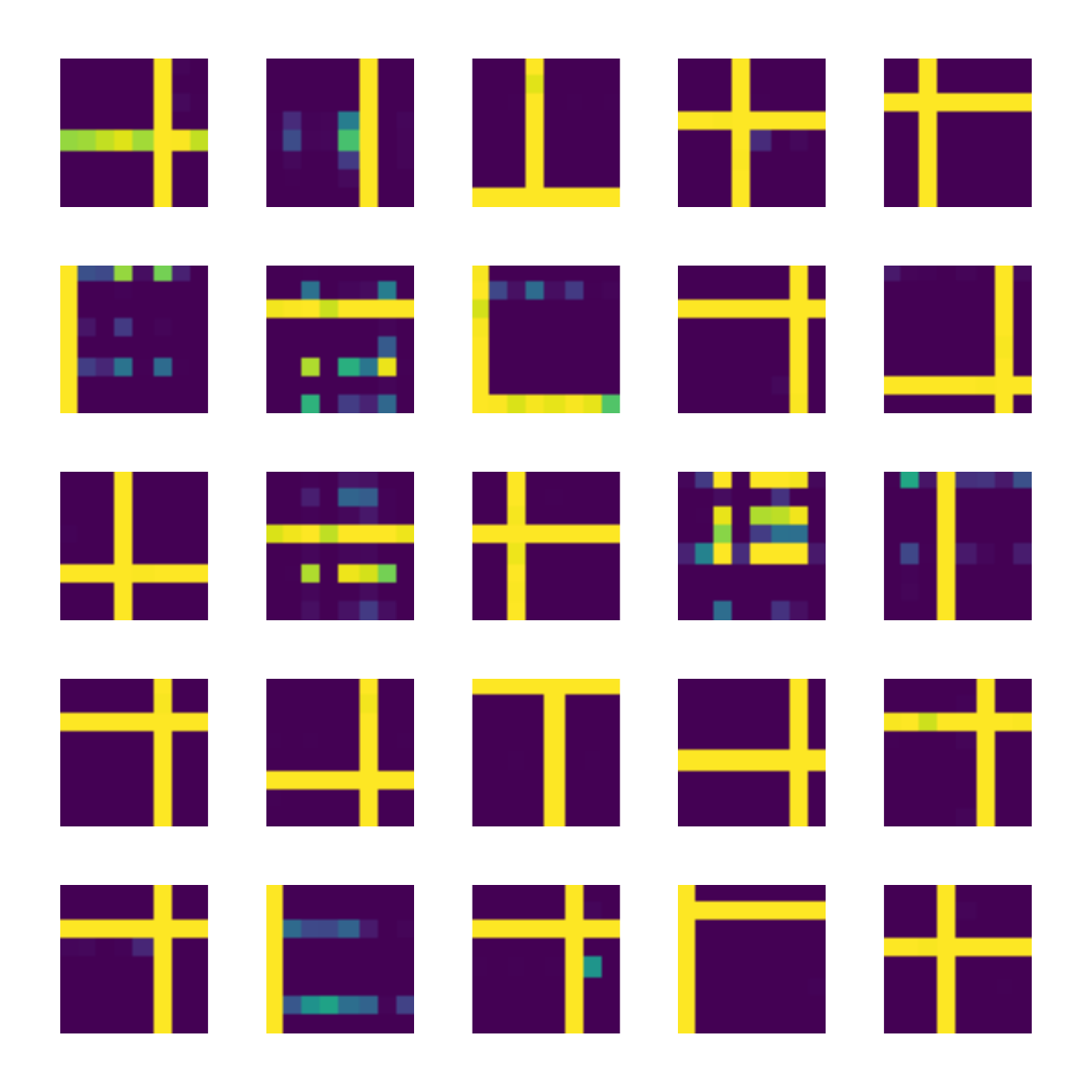}
         \caption{}
         \label{fig: vae_bar_gen}
     \end{subfigure}
     \hfill
     \begin{subfigure}[b]{0.32\textwidth}
         \centering
         \includegraphics[width=\textwidth]{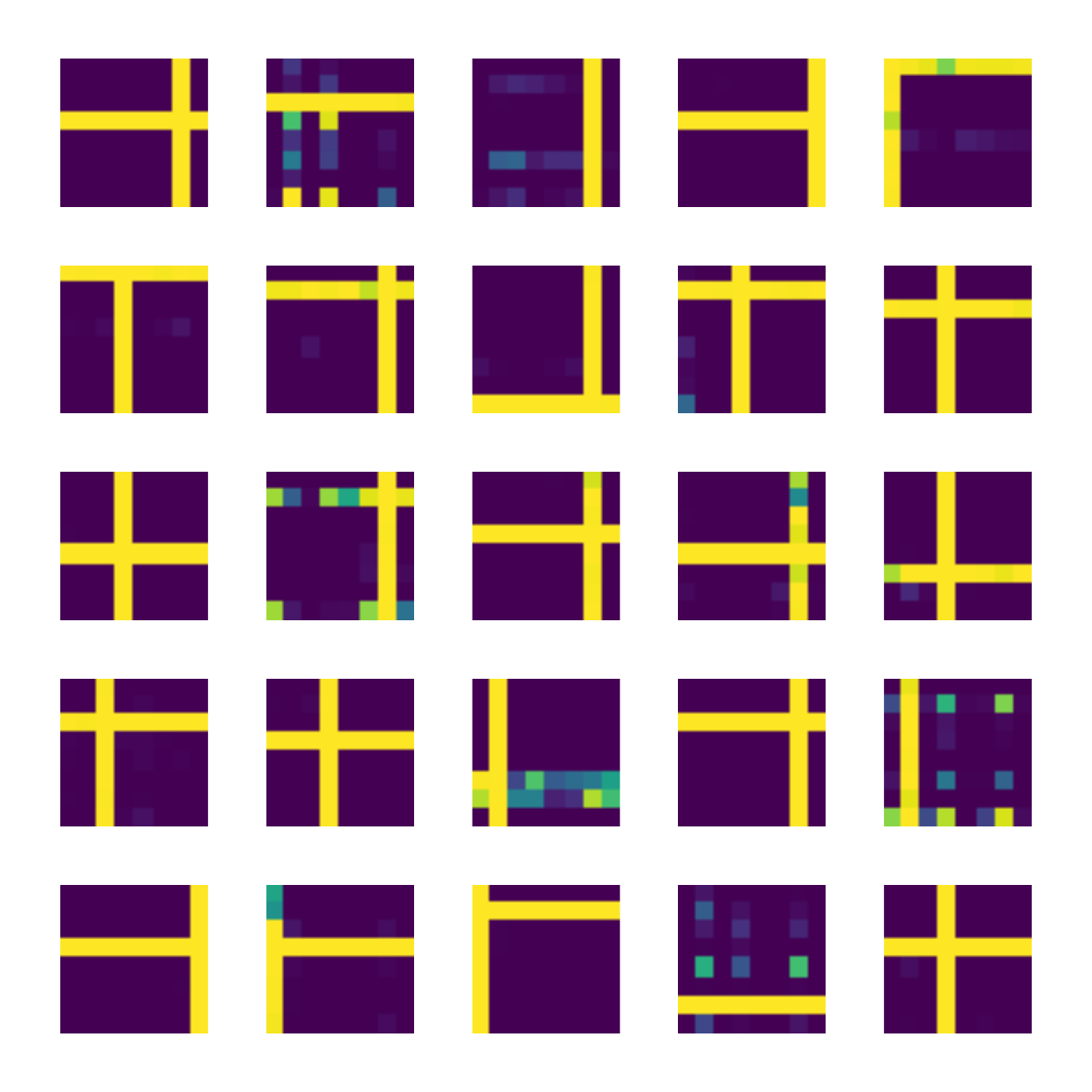}
         \caption{}
         \label{fig: svae_bar_gen}
     \end{subfigure}
     \hfill
     \begin{subfigure}[b]{0.32\textwidth}
         \centering
         \includegraphics[width=\textwidth]{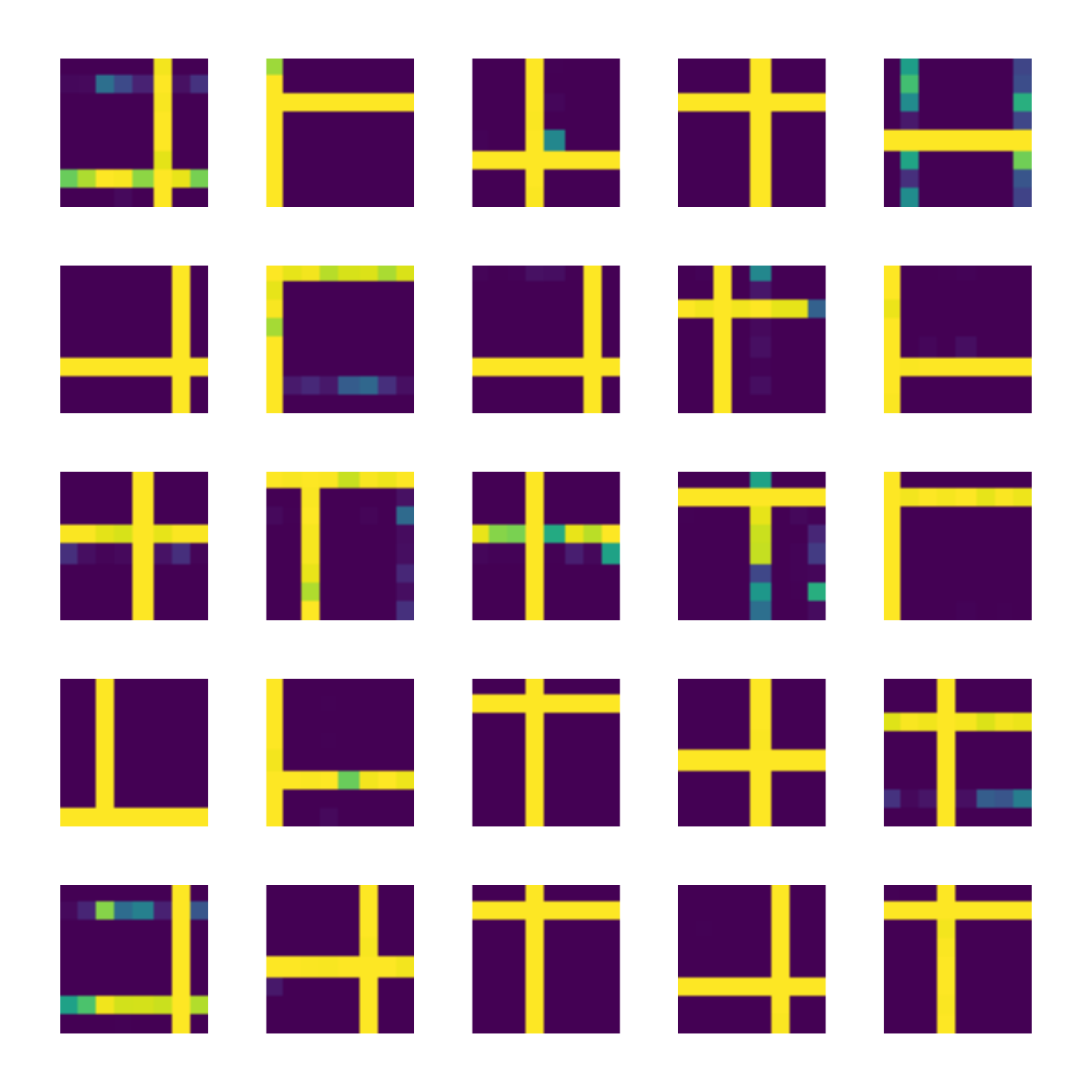}
         \caption{}
         \label{fig: TreeSRVAE_bar_gen}
     \end{subfigure}
        \caption{\textbf{Generation from trained models on the ``bar" dataset with ``side-depdent" prior.} (a) VAE (mean-field latent prior and mean-field recognition inference); (b) SVAE (tree-structured latent prior and mean-field recognition inference); (c) TreeSRVAE (tree-structured latent prior and tree-structured recognition inference). Note that all samples are not cherry-picked.
        }
        \label{fig: tree_generation_comparison}
\end{figure*}

Firstly, visual inspection of the randomly samples from the three trained models in Figure~\ref{fig: tree_generation_comparison} indicates TreeSRVAE generates more ``cross" patterns than SVAE and VAE. The qualitative indicates TreeSRVAE has learned a better generative model than the other two models.

 Now for the quantitative comparison. Given the relatively small latent state space ($16$ binary latents, leading to $2^{16}=65536$ possible latent configurations). We exhaustively generate samples from all possible latent configurations given the three models. For each generated sample, we compute the most similar ``cross" pattern (in terms of squared error) with respect to the sample, and their squared distance. We then compute the averaged smallest squared error given the generated samples of all possible $2^{16}$ latent configurations for the three models. The statistics is reported in Table~\ref{tab: avg_closest_cross}. Namely, for each $z\in\mathbb{R}^{16}$, 
\begin{equation}
\begin{split}
    \hat{y} &= p(y|z, \gamma)\\
    y^{\ast} &= \argmax_{y\in\mathcal{Y}}||y-\hat{y}||^2\\
    d_{z} &= ||y^{\ast}-\hat{y}||^2
\end{split}
\end{equation}
where $\mathcal{Y}$ denotes the set of all possible cross patterns ($64$ different patterns in an $8\times8$ square-grid. Hence we compare $\tilde{d} = \frac{1}{|\mathcal{Z}|}\sum_{z\in\mathcal{Z}}d_{z}$, where $\mathcal{Z}$ is the set of all possible latent configurations.

Hence we have shown that, despite having achieved similar free energy, TreeSRVAE enables stronger generation both qualitatively and quantitatively.

\begin{table}[h!]
    \centering
    \begin{tabular}{|c|c|c|c|}
    \hline
    & VAE & SVAE & TreeSRVAE  \\
    \hline
    $\tilde{d}$ & $1.297$ & $0.559$ & $\vec{0.351}$ \\
    \hline
    \end{tabular}
    \caption{Comparison of $\tilde{d}$ for VAE, SVAE, and TreeSRVAE.}
    \label{tab: avg_closest_cross}
\end{table}

\subsubsection{Tree-Structured Latent PGM with Gaussian Factors}

We note that the tree-structured recognition network generalises beyond discrete latents. Namely, we developed a Gaussian-TreeSRVAE model (details in appendix~\ref{sec: app_gaussian_factors}) and applied to the MNIST dataset~\cite{lecun2010mnist}. Table~\ref{tab: mnist_free_energy} show that Gaussian-TreeSRVAE outperforms the two baseline models. 

\begin{table}[h!]
    \centering
    \begin{tabular}{|c|c|c|c|}
    \hline
         & VAE & SVAE & Gaussian-TreeSRVAE  \\
         \hline
    Free energy     & $89.54\pm 0.21$ & $91.52\pm 0.41$ & $\vec{88.39 \pm 0.55}$ \\
    \hline
    \end{tabular}
    \caption{Free energy of VAE, SVAE and Gaussian-TreeSRVAE trained on MNIST dataset (given $100$ training epochs).}
    \label{tab: mnist_free_energy}
\end{table}



\subsection{SR-nlGPFA}
\label{sec: app_aea_svgpfa}

\subsection{Full Quantitative Evaluation Comparison}
The complete evaluation comparison between SR-nlGPFA and the selected baselines in terms of both the SMSE and the negative log-likeihood (NLL) is shown in Table~\ref{tab: synthetic_eeg_results}.


\subsubsection{Complexity Analysis of SR-nlGPFA}
\label{sec: app_complexity}
\begin{figure*}[h]
     \centering
     \begin{subfigure}[b]{0.43\textwidth}
         \centering
         \includegraphics[width=\textwidth]{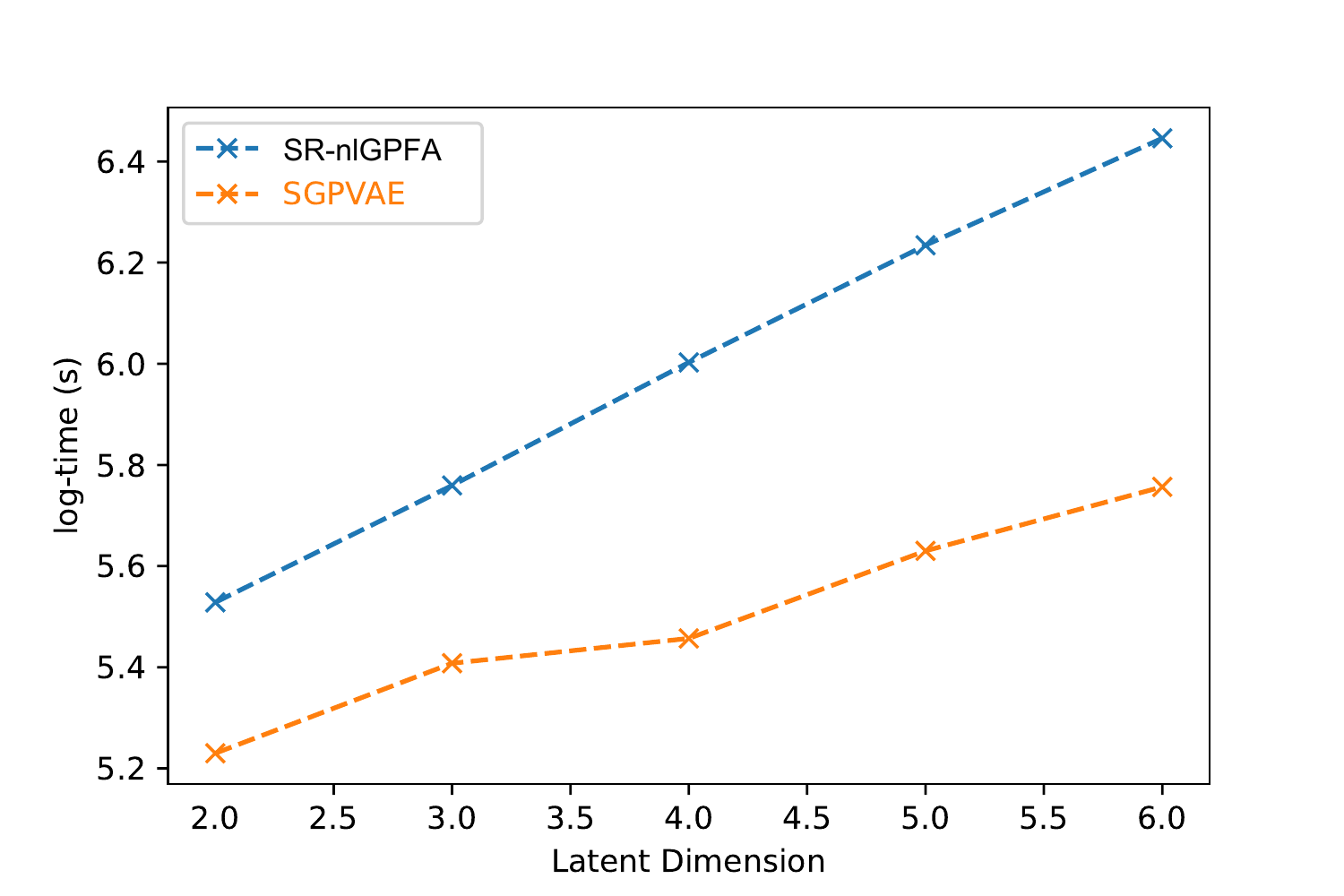}
         \caption{}
         \label{fig: eeg_time_LD}
     \end{subfigure}
     \hfill
     \begin{subfigure}[b]{0.43\textwidth}
         \centering
         \includegraphics[width=\textwidth]{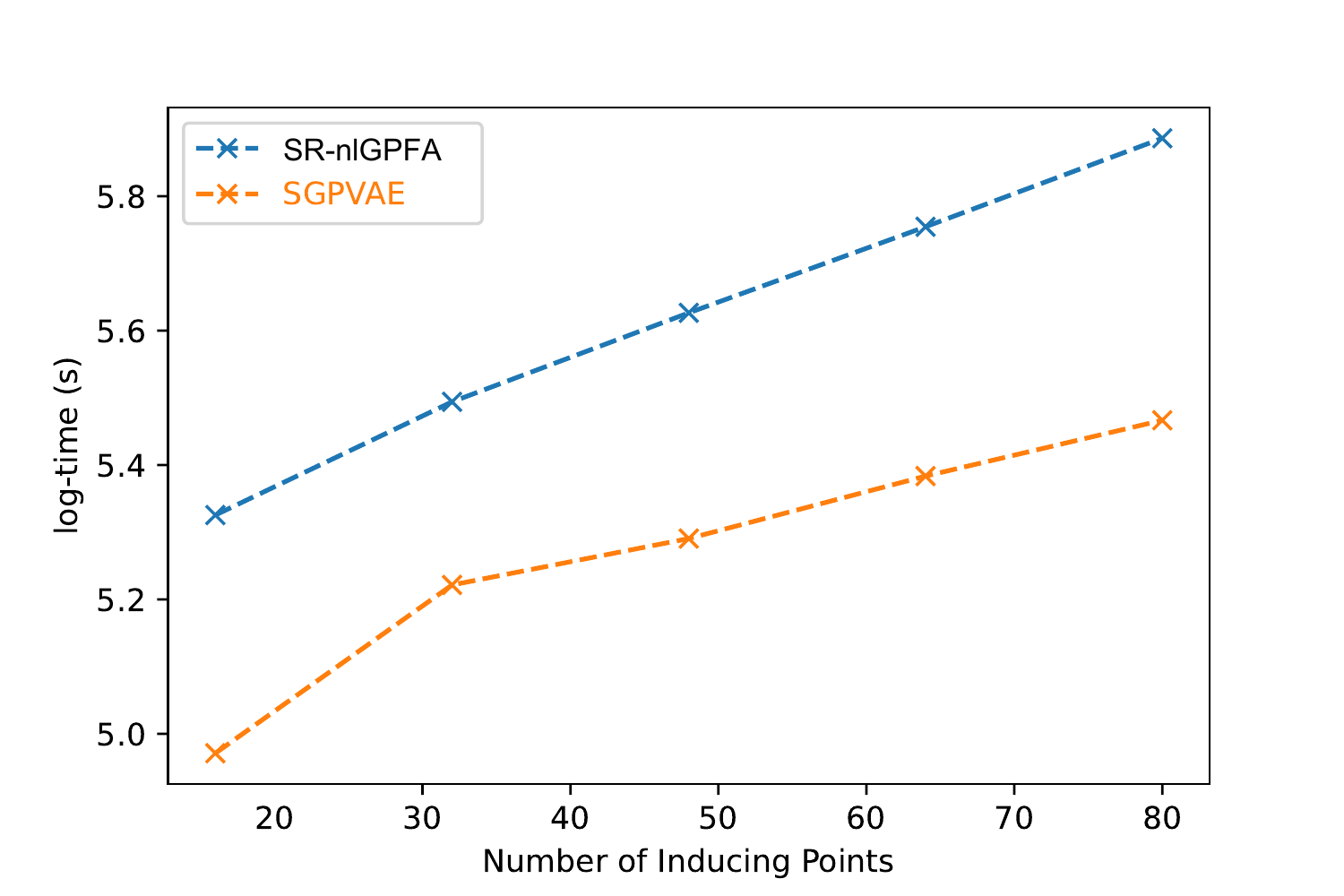}
         \caption{}
         \label{fig: eeg_time_NI}
     \end{subfigure}
        \caption{Computation time (log-scale) comparison between SR-nlGPFA and SGP-VAE with varying latent dimension (a) and number of inducing points per latent dimension (b).
        }
        \label{fig: eeg_time}
\end{figure*}

We note that SR-nlGPFA, despite allowing the inference of full-covariance structure over all latent processes, does not incur higher-order complexity than standard SVAEs with GP-latents (e.g., SGP-VAE~\cite{ashman2020sparse}), hence allowing scalable application. The major difference between SR-nlGPFA and SGP-VAE lie in the inference step, where SGP-VAE requires only inverting the covariance matrices for the latent-specific inducing points, and SR-nlGPFA needs to numerically invert the covariance matrices over all inducing points across latent dimensions. Hence the complexities for such operation is $\mathcal{O}(K^3\bar{M}^3)$ and $\mathcal{O}(K\bar{M}^3)$ for SR-nlGPFA and SGP-VAE, respectively. Hence SR-nlGPFA incurs additional computations by a factor of $K^2$. However, we note that we usually seek a low-dimensional GPFA latents that characterise the manifold upon which the high-dimensional neural trajectories lie within, hence $K$ is usually chosen to be small (e.g., $K=4$ for the single-cell spiking dataset from Section~\ref{sec: aea_svgpfa_neural} and $K=2$ for the EEG dataset considered in Section~\ref{sec: aea_svgpfa_eeg}). Hence overall speaking, we expect considerably small increase in computation time. We show the comparison between the running times with varying latent dimensions and number of inducing points for SR-nlGPFA and SGP-VAE in Figure~\ref{fig: eeg_time}, which is in accordance with our hypothesis that small additional computational cost is incurred with SR-nlGPFA.

\subsubsection{Further Analysis of SR-nlGPFA on Neural Population Spiking Dataset}
\label{sec: app_aea_neural}

\begin{figure}[h!]
    \centering
    \includegraphics[width=\linewidth]{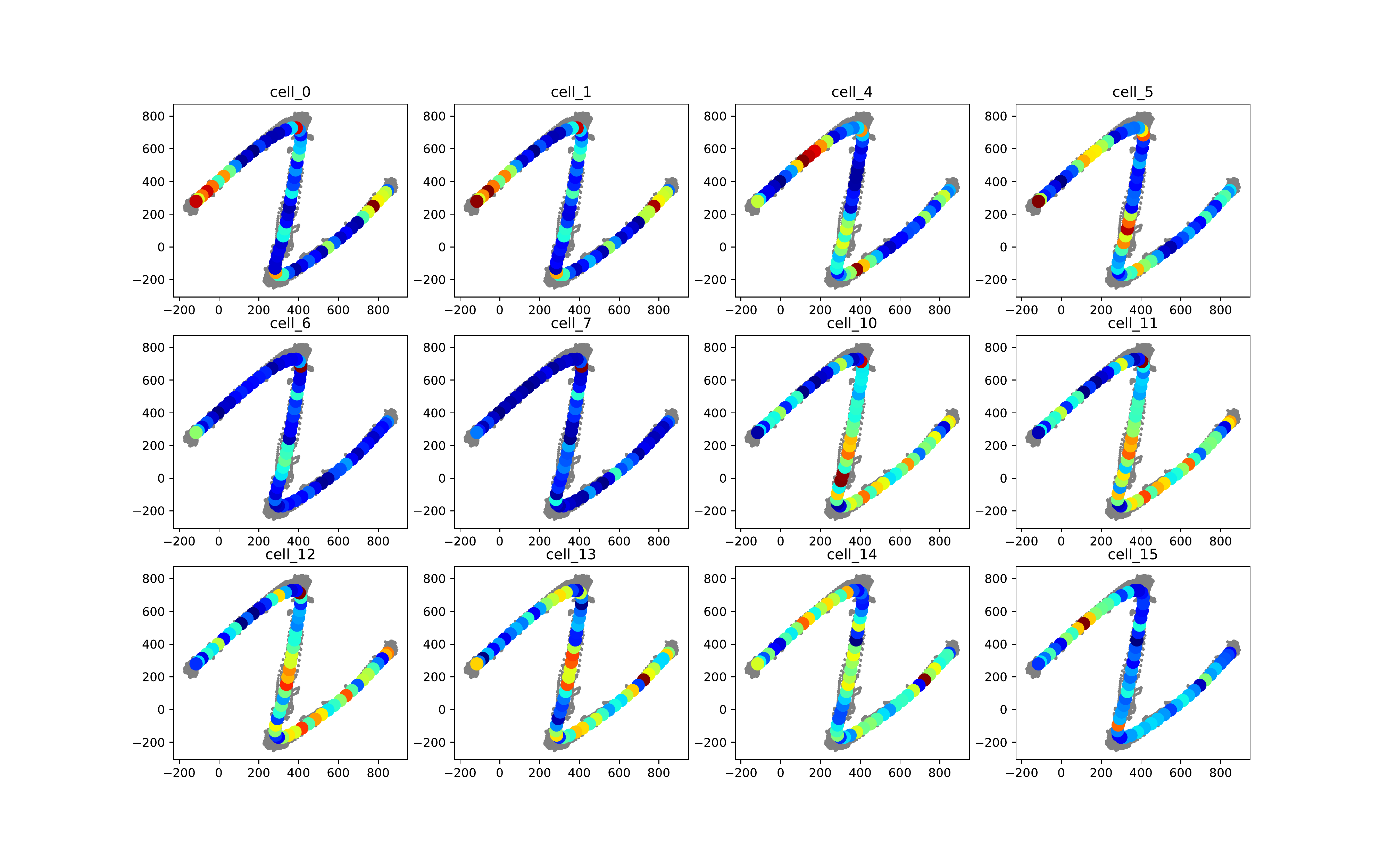}
    \caption{Firing patterns of selected CA1 and mEC neurons along the Z-shaped track (color represents firing rate).}
    \label{fig: exemplary_pc_fp}
\end{figure}

The firing fields of some exemplary neurons on the Z-shaped track in shown in Figure~\ref{fig: exemplary_pc_fp}. Visual inspections show that the majority of the recorded cells show spatial modulation, and resembling the firing patterns of place cells and (potentially) grid cells, amongst the recorded neurons there also exists ones whose firing patterns resemble that of interneurons (e.g., third neurons from the left on the third row).

\textbf{Extraction of behavioural-covariate-modulated neurons predicted by the trained SR-nlGPFA model. } 

Given the behavioural covariate of interests, we wish to predict the neurons whose firing pattern exhibit strongest modulation with respect to the behavioural covariate using the trained SR-nlGPFA model (in a completely unsupervised fashion).

Specifically, given the projection matrices from the CCA fitting, we could qualitatively identify the latent processes that are most correlated with the target behavioural covariate. Given the generative likelihood, we could compute the gradient of the output Poisson rate parameters for each neuron with respect to the selected latent variable (the Jacobian matrix) using AutoDiff as found in most deep learning libraries~\cite{NEURIPS2019_9015}. 
We define the \textit{relevance score} of a latent process, $f$, with respect to the firing of neuron $s$, $\Delta_{f}(s)$ as the average squared-norm of the gradient of the associated mapping (given the likelihood function, Eq.~\ref{eq: poisson_likelihood}).
\begin{equation}
    \Delta_{f}(s) = \int_{t=0}^{T} dt ||\frac{\partial \lambda_{\text{NN}}(\vec{f}(t))_{s}}{\partial f}||^2 \approx \frac{1}{N}\sum_{n=1}^{N}||\frac{\partial \lambda_{\text{NN}}(\vec{f}(x_{n}))_{s}}{\partial f}||^2
\label{eq: relevance_measure}
\end{equation}
We categorise the top $20\%$ of the neurons with the highest (or lowest if negatively correlated) relevance score as the model-predicted neurons that are modulated by the target behavioural covariate.

\textbf{Additional svGP Inference Step with Changed Inducing Locations}

We note that since the recognition network outputs local evidence potentials on $\vec h$, the computation of $q(\vec U)$ is not constrained to a single set of inducing locations $\vec Z$, rather we are free to choose the set of inducing points to work with~\eqref{eq: svgpfa_qu}. During training, due to the computational constraints of stochastic mini-batch training, it is impractical to use a large number of inducing points. However, the small number of inducing points leads to temporal chunking artifacts in the variational posterior distribution of the latent processes ($q(\vec f)$) over the entire sequence. Hence SR-nlGPFA enables an additional svGP inference step given the trained recognition network and an expanded set of inducing points, leading to smoother temporal interpolation, hence alleviating the temporal chunking effects. 

\textbf{Quantitative Analysis of SR-nlGPFA across Experimental Sessions}

\begin{figure*}[t]
     \centering
     \begin{subfigure}[b]{0.33\textwidth}
         \centering
         \includegraphics[width=\textwidth]{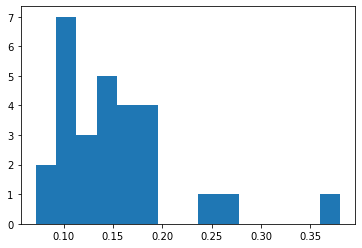}
         \caption{}
         \label{fig: off_diagonal_ratio}
     \end{subfigure}
     \hfill
     \begin{subfigure}[b]{0.6\textwidth}
         \centering
         \includegraphics[width=\textwidth]{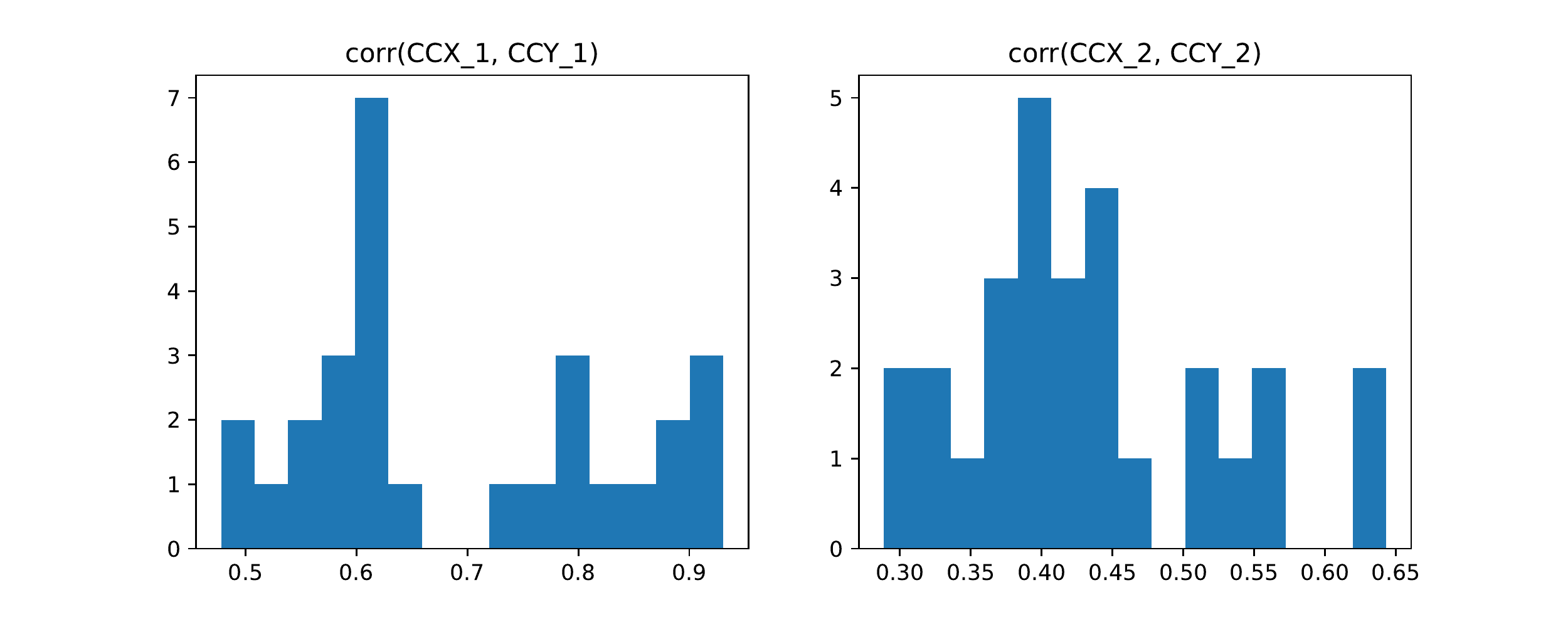}
         \caption{}
         \label{fig: cc_corr}
     \end{subfigure}
        \caption{Quantitative analysis of SR-nlGPFA across experimental sessions. (a) Histogram of
          the ratio of the magnitude of the off-diagonal entries relative to that of all entries in
          the covariance matrix of $q(\vec{f})$; (b) Histogram of correlation coefficients between $CCX\{1, 2\}$ and $CCY\{1, 2\}$.
        }
        \label{fig: quantitative_all}
\end{figure*}

Here we quantitatively evaluate various aspects of the posterior latent GPs given the trained models. We train independent models of SR-nlGPFA with the same architectures and training procedures (see appendix~\ref{sec: app_implementation_detail} for implementation details) on the single-cell population spiking datasets from each of the $28$ experimental sessions considered. Firstly, we examine if full-covariance structure is a necessary assumption. We compute the ratio between the magnitudes of the off-diagonal entries and that of all entries in the posterior covariance matrix ($\vec{S}^{f}_{n}$, for all $n$, Section~\ref{sec: aea_svgpfa_neural}), $\sum_{n}\frac{\sum_{i, j, i\neq j}|\vec{S}_{n, ij}^{f}|}{\sum_{i, j}|\vec{S}_{n, ij}^{f}|}$. From Figure~\ref{fig: off_diagonal_ratio} we observe that the posterior latent covariance matrices of all sessions have non-trivial off-diagonal elements, indicating the existence of posterior correlations induced by ``explaining away" and the necessity of posterior inference with full covariance structure for capturing the induced correlations.

In order to assess the quality of the learned posterior latents, we perform two-dimensional Canonical Correlation Analysis (CCA, \cite{hardoon2004canonical}) on the learned posterior means and the behavioural covariates. In Figure~\ref{fig: cc_corr} we observe that the correlations between the extract canonical correlates (CC) of the posterior means and the behavioural covariates exhibits high correlation over all sessions, hence indicating the extracted posterior latent dimensions captures large proportion of the information in the behavioural variables.

\begin{figure}
    \centering
    \vspace{-60pt}
    \includegraphics[angle=90,origin=c, width=.8\textwidth]{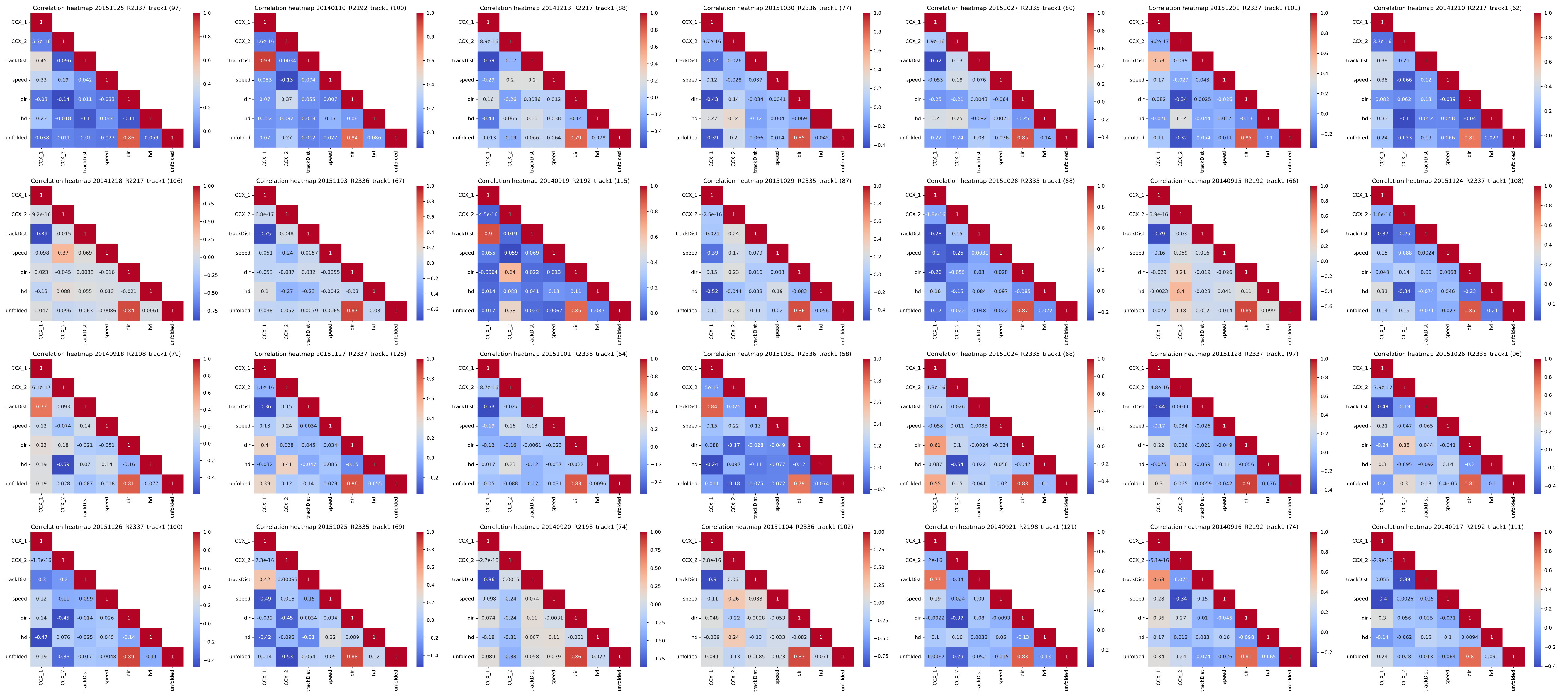}
    \caption{Heatmaps of correlation coefficients between $CCX\{1, 2\}$ of posterior latent mean given SR-nlGPFA and the behavioural covariates.}
    \label{fig: cca_heatmap_all}
\end{figure}

We further show the correlations between the $CCX\{1, 2\}$ (CCs of the latent posterior means) and each of the behavioural covariates for all sessions in Figure~\ref{fig: cca_heatmap_all}. We observe strong correlations between one or both of the CCXs with the (unfolded) distance along the track quantity, showing that the extracted latent variables given the population spiking data is strongly indicative of the spatial location of the rat along the Z-shaped track, and corresponds nicely with prior knowledge that the majority of the recorded cells are located in the CA1 region of hippocampus and are well-known to exhibit significant spatial modulation~\cite{o1978hippocampus}. Moreover, in most sessions we observe high correlations (in terms of magnitude) between the CCXs and the direction of travelling, again conforming with experimental evidence that CA1 place cells firing are modulated by direction along the linear track (the Z-shaped track environment is usually interpreted as a 1D linear track rather than a 2D environment~\cite{o1978hippocampus}). We additionally observe that the CCXs of the learned latent variables in most sessions show strong correlation with respect to speed of travelling, which also has nice experimental correspondence~\cite{mcnaughton1983contributions, kropff2015speed}.

\textbf{Comparison with SGP-VAE}

\begin{figure*}[h]
     \centering
     \hspace{-60pt}
     \begin{subfigure}[b]{0.46\textwidth}
         \centering
         \includegraphics[width=1.2\textwidth]{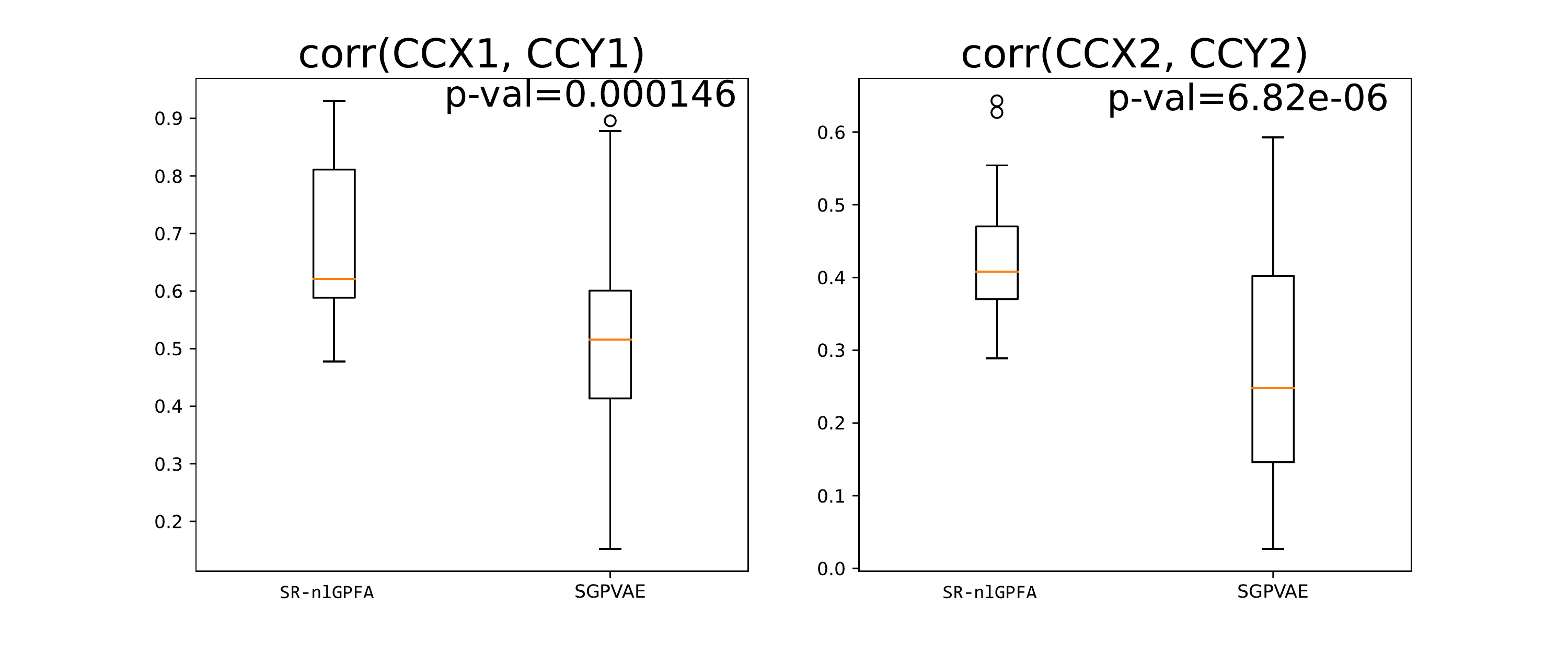}
         \caption{}
         \label{fig: CCX_corr_comparison_all}
     \end{subfigure}
    \qquad
     \begin{subfigure}[b]{0.46\textwidth}
         \centering
         \includegraphics[width=1.3\textwidth]{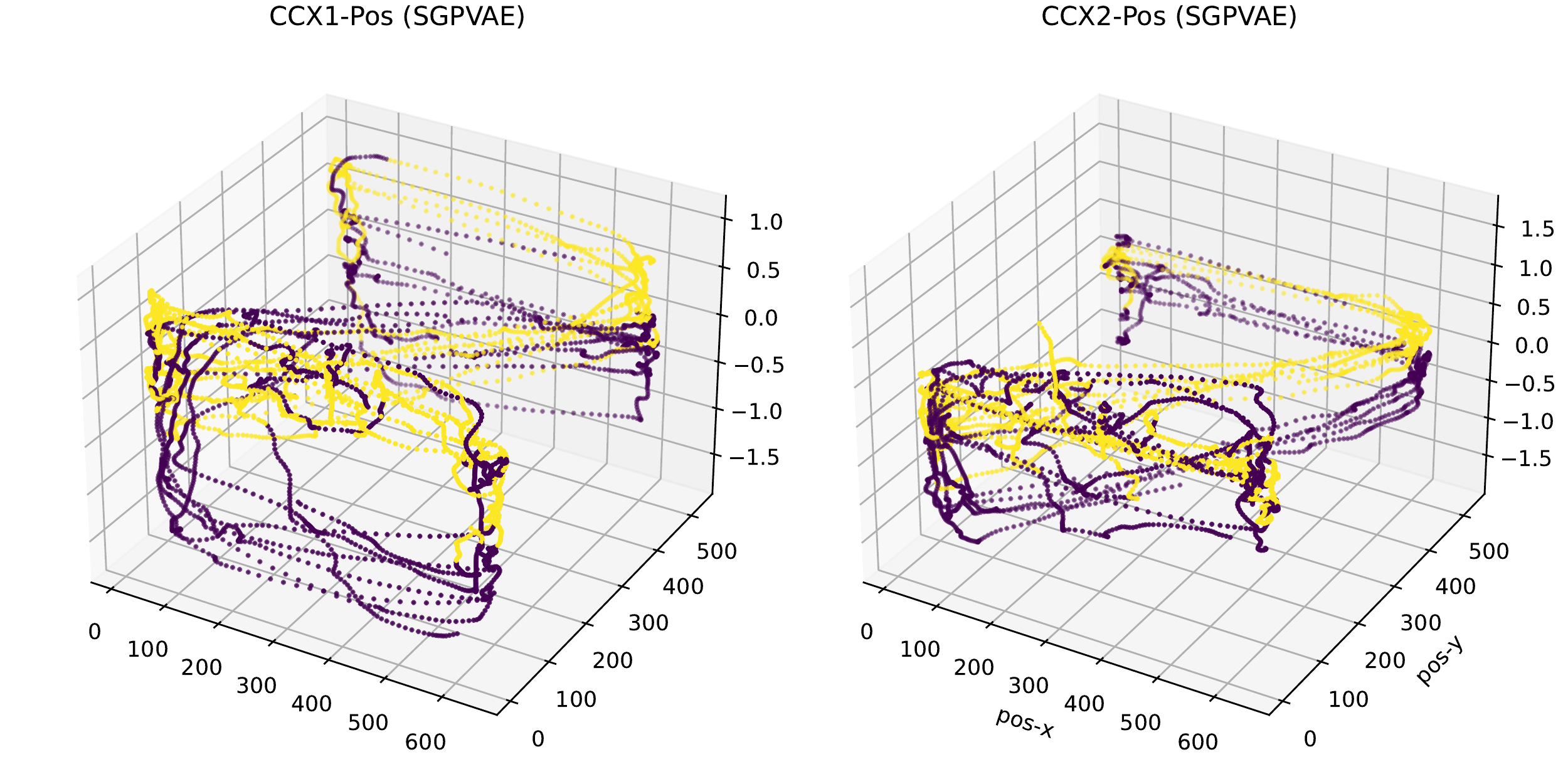}
         \caption{}
         \label{fig: cc_pos_orig}
     \end{subfigure}
        \caption{Comparison with SGP-VAE. (a) Comparison of $corr(CCX1, CCY1)$ (left) and $corr(CCX2, CCY2)$ (right) between SR-nlGPFA and SGP-VAE; (b) posterior mean of the latent process learned by SGP-VAE against x- and y-location of the rat, color indicating direction of travelling (yellow: inbound, magenta: outbound).
        }
        \label{fig: comparison_with_SGP-VAE}
\end{figure*}

By applying SGP-VAE to the data from the same session studied in Section~\ref{sec: aea_svgpfa_neural}, we perform similar CCA analysis on the learned latent variables of SGP-VAE. We observe that the correlations between $CCX\{1, 2\}$ and $CCY\{1, 2\}$ of SR-nlGPFA are significantly larger than that of SGP-VAE (with standard t-test, p-values shown in the figure; Figure~\ref{fig: CCX_corr_comparison_all}). 

We additionally examine the same session we looked at in Section~\ref{sec: aea_svgpfa_neural}, and plot the $CCX\{1, 2\}$ of the learned latent variables of SGP-VAE against the x- and y-positions. By comparing to the similar plot with SR-nlGPFA (Figure~\ref{fig: cc_pos_aea}), we observe that the resulting plot does not show clear clustered structure with respect to spatial and direction covariates compared to that of SR-nlGPFA, hence illustrating that SR-nlGPFA learns qualitatively better latent dimensions than SGP-VAE with respect to the underlying behavioural covariates.

\section{Implementation Details}
\label{sec: app_implementation_detail}

\textbf{``Bar" Dataset}

The details of the generation of the ``Bar" dataset can be found in appendix~\ref{sec: bar_test}. We use the same neural network architectures for all models considered (VAE, SVAE, TreeSRVAE): both the recognition and generative networks are MLPs with two hidden layers of $50$ hidden units, with ReLU non-linear activation function, and the latent dimension is $16$. All models and sessions are trained with Adam optimiser with learning rate of $5\times 10^{-4}$ over $3500$ epochs with a batch-size of $256$~\cite{kingma2014adam}. 

\textbf{Neural Population Spiking Dataset}

We use the same neural network architectures for both SR-nlGPFA and SGP-VAE: both the recognition and generative networks are MLPs with two hidden layers with $256$ hidden units, with ReLU activatieon function. The latent dimensions for both model is $6$, which we choose via cross-validation. Note that the latent dimension of $6$ corresponds to the results from other existing unsupervised latent feature extraction works on CA1 neuron firing patterns (e.g., \citet{nieh2021geometry} reported $5-7$ is optimal for the latent dimension with their manifold inference model). The dimension of the neurla feature ($h$) in the GPFA generative model (Eq.~\ref{eq: gpfa_gen}) is $20$. Both models are trained with Adam optimiser with learning rate of $1\times 10^{-4}$ over $400$ epochs with a batch-size of $128$~\cite{kingma2014adam}. The number of inducing points for each latent dimension is $64$ for both models. 

For the additional SVGP inference step after training with SR-nlGPFA (SGP-VAE does not allow the additional SVGP inference step as discussed in Section~\ref{sec: aea_svgpfa_neural}), we set the number of inducing points to be $1000$ for all latent dimensions to smooth out the temporal chunking artifacts caused by the small number of inducing points in training due to scalability concerns with stochastic optimisation.

\end{document}